\documentclass{article} 
\usepackage{iclr2026_conference,times}

\usepackage{amsmath,amsfonts,bm}

\def\eqref#1{equation~\ref{#1}}

\def\1{\bm{1}}

\DeclareMathAlphabet{\mathsfit}{\encodingdefault}{\sfdefault}{m}{sl}
\SetMathAlphabet{\mathsfit}{bold}{\encodingdefault}{\sfdefault}{bx}{n}

\usepackage[utf8]{inputenc} 
\usepackage[T1]{fontenc}    
\usepackage{hyperref}
\usepackage{url}
\usepackage{float}
\usepackage{booktabs}       
\usepackage{amsfonts}       
\usepackage{nicefrac}       
\usepackage{microtype}      
\usepackage[svgnames,dvipsnames,x11names]{xcolor}
\usepackage{amsmath}
\usepackage{amssymb}
\usepackage{mathtools}
\usepackage{amsthm}
\usepackage{bbm}
\usepackage{graphicx}
\usepackage{subcaption}
\usepackage{placeins}
\usepackage{tikz}
\usepackage{caption}
\usepackage{makecell}
\usepackage{hhline}
\usetikzlibrary{shadows}

\title{Oh-A-DINO: Understanding and Enhancing Attribute-Level Information in Self-Supervised Object-Centric Representations}

\author{Stefan Sylvius Wagner \\
Department of Computer Science\\
Heinrich Heine University\\
\And
Stefan Harmeling \\
Department of Computer Science \\
Technical University of Dortmund \\
}

\makeatletter

\renewcommand{\@maketitle}{\vbox{\hsize\textwidth
{\LARGE\sc \@title\par}
\lhead{} 
\def\And{\end{tabular}\hfil\linebreak[0]\hfil
        \begin{tabular}[t]{l}\bf\rule{\z@}{24pt}\ignorespaces}%
\def\AND{\end{tabular}\hfil\linebreak[4]\hfil
        \begin{tabular}[t]{l}\bf\rule{\z@}{24pt}\ignorespaces}%
\begin{tabular}[t]{l}\bf\rule{\z@}{24pt}\@author\end{tabular}%
\vskip 0.3in minus 0.1in}}
\makeatother

\begin{document}

\maketitle

\begin{abstract}
  Object-centric understanding is fundamental to human vision and required for complex reasoning. 
  Traditional methods define slot-based bottlenecks to learn object properties explicitly, while recent self-supervised vision models like DINO have shown emergent object understanding. 
  We investigate the effectiveness of self-supervised representations from models such as CLIP, DINOv2 and DINOv3, as well as slot-based approaches, for multi-object instance retrieval, 
  where specific objects must be faithfully identified in a scene. 
  This scenario is increasingly relevant as pre-trained representations are deployed in downstream tasks, e.g., retrieval, manipulation, and goal-conditioned policies that demand fine-grained object understanding.   
  Our findings reveal that self-supervised vision models and slot-based representations excel at identifying edge-derived geometry (shape, size) but fail to preserve non-geometric surface-level cues (colour, material, texture), which are critical for disambiguating objects when reasoning about or selecting them in such tasks.
 We show that learning an auxiliary latent space over segmented patches, where VAE regularisation enforces compact, disentangled object-centric representations, 
 recovers these missing attributes. 
 Augmenting the self-supervised methods with such latents improves retrieval across all attributes, 
 suggesting a promising direction for making self-supervised representations more reliable in downstream tasks that require precise object-level reasoning.
\end{abstract}

\section{Introduction}
\label{sec:intro}
Object-centric understanding is central to human vision and a prerequisite for complex reasoning \citep{vanSteenkiste2019Objects,schoelkopf2021causal}. Learning such representations remains challenging: slot-based methods explicitly disentangle object properties \citep{locatello2020slotattention,wu2023slotdiffusion}, but may compromise global scene understanding \citep{montero2024successes}, while large self-supervised models such as DINO and DINOv2 \citep{caron2021dino,oquab2023DINOv2} show emergent object structure without task-specific supervision. This makes them attractive as general-purpose vision backbones, increasingly used in downstream tasks that require object-level reasoning \citep{uelwer2023survey,seitzer2023bridging,zhou2024dinowm,wagner2024jci,mazoure2023embodied}.

Many of these downstream tasks such as retrieval, embodied manipulation, or goal-conditioned policies depend on representations that can reliably capture the properties needed to disambiguate between multiple objects in a scene. Whether current pre-trained and object-centric representations provide this level of fidelity, however, remains unclear.

We ask: \emph{Do existing representations preserve the attribute-relevant information needed to reliably disambiguate objects in multi-object settings?}

Our findings reveal that while self-supervised vision models and slot-based representations excel at capturing geometric structure such as shape and size, they fail to preserve non-geometric surface-level cues such as colour and material. This leads to systematic errors when disambiguating otherwise similar objects.
To address this, we propose \emph{Object-Aware-DINO (Oh-A-DINO)} which augments self-supervised representations from DINOv2-S with object-centric latent vectors learned from segmented patches using a Variational Autoencoder \citep{kingma2019introduction} (VAE). The VAE’s regularisation encourages a compact latent space that captures these fine-grained properties, and concatenating the latents with global features improves alignment across all object attributes.

We demonstrate that this approach improves multi-object instance retrieval on CLEVR \citep{johnson2017clevr} and CLEVRTex \citep{karazija2021clevrtex}, with particularly notable gains in colour and material matching. 
We also show that these improvements transfer to real-world instance retrieval settings on the Stanford Cars \citep{krause2013stanford} dataset. 
Overall, this highlights a broader challenge: widely used pre-trained representations, though powerful, may miss the kinds of attribute-level information that downstream tasks need for accurate object-level reasoning.
\paragraph{Our contributions are threefold:}
(i) We identify a key limitation of both self-supervised and slot-based representations: they capture geometric attributes well but fail to preserve non-geometric object attributes such as colour and material, which are critical for disambiguating objects in multi-object scenes.  

(ii) We propose a simple and modular method to augment self-supervised representations with object-centric latents learned from a VAE, without retraining the backbone.  

(iii) We demonstrate that this approach improves retrieval across all attributes, providing more reliable object-level representations and suggesting a promising direction for enhancing pre-trained representations in downstream tasks that require detailed object reasoning.

\FloatBarrier
\begin{figure}[htbp!]
  \centering
  \begin{tikzpicture}
  \node at (-4,1.5) {\includegraphics[width=0.99\linewidth]{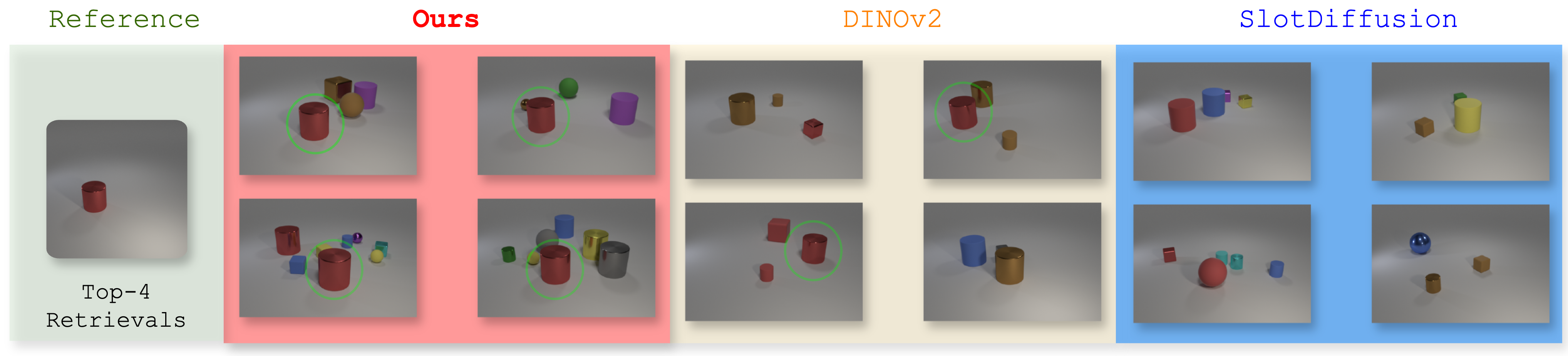}};

  \node[align=left] (bg) at  (-0.3,-1.55) {\tiny\fontfamily{pcr}\selectfont \textbf{Legend:} S: Shape, D: Size, M: Material, C: Colour};
  \node[fill=brown!5, minimum width=13.78cm, minimum height=1.2cm] (bg) at (-3.97,-0.7) {};

    \node[align=left] (bg) at  (-9.87,-1.45) {\scriptsize\fontfamily{pcr}\selectfont \scalebox{0.6}{\textcolor{gray}{Top-10}}\;\scriptsize\fontfamily{pcr}\selectfont \scalebox{0.6}{\textcolor{gray}{Precision(\%)}}};
    \node[align=left] (bg) at  (-9.9,-0.45) {\scriptsize\fontfamily{pcr}\selectfont \scalebox{0.7}{Single Attribute}};
    \node[align=left] (bg) at (-9.9,-0.9) {\scriptsize\fontfamily{pcr}\selectfont \scalebox{0.7}{\makecell{Multi Attribute \\  $\mathcal{X} = \{S,D,M\}$}}};
    \node[align=left] (bg) at (1.9,-1.35) {\scriptsize\fontfamily{pcr}\selectfont \scalebox{0.7}{\makecell{\textcolor{gray}{$\rightarrow$ More attributes}}}};

  \node[fill=red!15, minimum width=3.95cm, minimum height=1.15cm] (bg) at (-7.0,-0.7) {};
  \node[fill=yellow!15, minimum width=3.95cm, minimum height=1.15cm] (bg) at (-3.05,-0.7) {};
  \node[fill=cyan!10, minimum width=3.95cm, minimum height=1.15cm] (bg) at (0.9,-0.7) {};

  \node at (-2.05, -0.7){
  \begin{minipage}{1.0\textwidth}
    {\fontfamily{pcr}\selectfont \scriptsize
    \scalebox{0.73}{
    \begin{tabular}{|c|c|c|c|}
    \cline{1-4}
        \emph{S} & \emph{D} & \emph{M}& \emph{C} \\ \cline{1-4}
        94.7 & \textbf{96.7} & \textbf{98.6} & \textbf{96.6} \\ \cline{1-4}
        \hhline{====}  
        \hhline{====}  
        \emph{$\mathcal{P}_1(\mathcal{X})$} & \emph{$\mathcal{P}_2(\mathcal{X})$} & \emph{$\mathcal{P}_3(\mathcal{X})$} & \emph{$\mathcal{P}_3(\mathcal{X})$+$C$}\\ \cline{1-4}
    \textbf{96.6} & \textbf{87.1}& \textbf{72.8} & \textbf{56.4}\\ \cline{1-4}
    \end{tabular}}}
    \label{tab:single-single}
  \end{minipage}};
  \node at (1.9, -0.7){
    \begin{minipage}{1.0\textwidth}
      {\fontfamily{pcr}\selectfont \scriptsize
      \scalebox{0.73}{
        \begin{tabular}{|c|c|c|c|c|}
      \cline{1-4} 
      \emph{S} & \emph{D} & \emph{M}& \emph{C} \\ \cline{1-4}
            \textbf{95.4} & 81.2 & 96.1 & 40.8 \\ \cline{1-4}
            \hhline{====}  
            \hhline{====}  
            \emph{$\mathcal{P}_1(\mathcal{X})$} & \emph{$\mathcal{P}_2(\mathcal{X})$} & \emph{$\mathcal{P}_3(\mathcal{X})$} & \emph{$\mathcal{P}_3(\mathcal{X})$+$C$}\\ \cline{1-4}
      90.9 & 76.2 & 62.8 & 13.0 \\ \cline{1-4}
      \end{tabular}}}
      \label{tab:single-single}
    \end{minipage}};
    \node at (5.85, -0.7){
      \begin{minipage}{1.0\textwidth}
        {\fontfamily{pcr}\selectfont \scriptsize
        \scalebox{0.73}{
            \begin{tabular}{|c|c|c|c|}
                \cline{1-4} 
                \emph{S} & \emph{D} & \emph{M}& \emph{C} \\ \cline{1-4}
                76.3 & 89.3 & 94.1 & 63.5 \\ \cline{1-4}
                \hhline{====}
                \hhline{====}
                \emph{$\mathcal{P}_1(\mathcal{X})$} & \emph{$\mathcal{P}_2(\mathcal{X})$} & \emph{$\mathcal{P}_3(\mathcal{X})$} & \emph{$\mathcal{P}_3(\mathcal{X})$+$C$} \\ \cline{1-4}
                82.2 & 61.4 & 39.5 & 12.0 \\ \cline{1-4}
                \end{tabular}}}
        
        \label{tab:single-single}
      \end{minipage}};
\end{tikzpicture}
  \caption{\textbf{DINOv2 representations and slot-based representations struggle at multi-object instance retrieval, due to weak object-specific features (largely colour) in its embeddings. 
  Our method is able to retrieve more relevant images by combining general scene and local representations.}
  DINOv2 representations excel at retrieving images where multiple attributes match such as shape and size, while retrieval degrades when adding colour.
  Slot-based representations lack specialisation, where retrievals sometimes match fine-grained features such as colour but often failing to retrieve a similar object altogether. 
  Our method performs the best being able to augment the DINOv2 representation to mitigate its shortcomings.}  
  \label{fig:intro}
\end{figure}
\FloatBarrier
\begin{figure}[htbp!]
  \captionsetup{name=Table, type=table}
  \caption{\textbf{Object-level features improve prediction for all attributes, while also improving retrieval of multiple simultaneous attributes.} The main improvement is 
  achieved on the colour(texture) attribute, which indicates that our method mitigates for the lacking learning signal in the other methods' representations. 
  Slot-based methods perform worse than the pretrained DINOv2 representations when predicting multiple attributes simultaneously.}
  \centering
  \begin{tikzpicture}

      \node[align=left] (bg) at  (-3.25,1.55) {{\fontfamily{pcr}\selectfont\tiny Multi-Attribute Retrieval}};
      \node[align=left] (bg) at  (-8.5,1.55) {{\fontfamily{pcr}\selectfont\tiny Single-Attribute Retrieval}};
      
      
      \node[fill=brown!5, minimum width=1.8cm, minimum height=2.4cm, rounded corners=0mm] (bg) at (-2.0,3.0) {};
      \node[fill=brown!5, minimum width=2.59cm, minimum height=2.4cm, rounded corners=0mm] (bg) at (-6.85,3.0) {};
      


      \node at (-7.2,3) {

      \begin{minipage}{1.0\textwidth}            
          {\fontfamily{pcr}\selectfont \footnotesize
          \scalebox{0.57}{
          \setlength{\arrayrulewidth}{0.7pt}  
          \setlength{\tabcolsep}{3pt}  

          \hspace*{0.3cm}
          \begin{tabular}{c|cccc|ccc!{\vrule width 2pt}ccc|cc|}
          \cline{2-13}
          & \multicolumn{4}{c|}{\rule[-1.2ex]{0pt}{4.1ex}CLEVR} & \multicolumn{3}{c!{\vrule width 2pt}}{CLEVRTex} & \multicolumn{3}{c|}{CLEVR} & \multicolumn{2}{c|}{CLEVRTex} \\ \cline{2-13}
          \multicolumn{1}{c|} {\makecell{\textbf{Attrib.\;Ablation} \\ \tiny{Top-10 Precision (\%)}}}  & \rule[-1.7ex]{0pt}{4.6ex} \emph{Shape} & \emph{Size} & \emph{Mat.}& \emph{Col.} & \emph{Shape} & \emph{Size} & \emph{Mat.}&  \emph{$\mathcal{P}_2(\mathcal{X})$} & \emph{$\mathcal{P}_3(\mathcal{X})$} & \emph{$\mathcal{P}_3(\mathcal{X})$+$C$} & \emph{$\mathcal{P}_2(\mathcal{X})$} & \emph{$\mathcal{P}_2(\mathcal{X})$+$C$} \\ \cline{1-13}
          \multicolumn{1}{|l|}{\rule[-1.7ex]{0pt}{4.6ex}\textcolor{DarkRed}{Oh-A-DINOv2\textbf{(Ours)}}} & 94.7{\tiny $\pm 2.0$} & \textbf{96.7}{\tiny $\pm 2.2$} & \textbf{98.6}{\tiny $\pm 0.0$} & \textbf{96.6}{\tiny $\pm 2.6$} & \textbf{90.7}{\tiny $\pm 2.1$}  & 85.3{\tiny $\pm 1.5$} & \textbf{45.8{\tiny $\pm 3.9$}}  & \textbf{85.2{\tiny $\pm 2.1$}} & \textbf{72.8{\tiny $\pm 1.4$}}& \textbf{56.4{\tiny $\pm 1.6$}}  & \textbf{56.0}{\tiny $\pm 0.2$} & \textbf{20.3}{\tiny $\pm 2.6$} \\ 

          \multicolumn{1}{|l|}{\rule[-1.7ex]{0pt}{0ex}\textcolor{DarkGoldenrod}{DINOv2}}     & 95.4{\tiny $\pm 1.1$} & 81.2{\tiny $\pm 1.2$} & 96.1{\tiny $\pm 1.0$}  & 40.8{\tiny $\pm 2.1$} & 87.3{\tiny $\pm 3.8$} & 80.3{\tiny $\pm 2.6$} & 24.2{\tiny $\pm 1.9$}  & 76.2{\tiny $\pm 3.1$}  & 62.8{\tiny $\pm 1.2$} & 13.0{\tiny $\pm 1.1$}  & 54.0{\tiny $\pm 0.1$}  & 12.2{\tiny $\pm 2.1$} \\ 
          \multicolumn{1}{|l|}{\rule[-1.7ex]{0pt}{0ex}\textcolor{Purple}{DINOv3}}     & \textbf{96.9}{\tiny $\pm 1.3$} & 85.5{\tiny $\pm 1.5$} & 97.0{\tiny $\pm 1.2$}  & 44.5{\tiny $\pm 1.9$} & 81.3{\tiny $\pm 0.7$} & \textbf{86.8}{\tiny $\pm 1.2$} & 11.8{\tiny $\pm 1.1$}  & 80.1{\tiny $\pm 1.2$}  & 64.8{\tiny $\pm 0.6$} & 13.7{\tiny $\pm 1.2$}  & 44.5{\tiny $\pm 0.2$}  & 4.2{\tiny $\pm 1.5$} \\ 
          \multicolumn{1}{|l|}{\rule[-1.7ex]{0pt}{0ex}\textcolor{DarkGreen}{CLIP}}     & 83.8{\tiny $\pm 2.3$} & 83.2{\tiny $\pm 1.8$} & 98.3{\tiny $\pm 0.2$}  & 66.8{\tiny $\pm 1.7$} & 81.4{\tiny $\pm 1.1$} & \textbf{90.1}{\tiny $\pm 1.4$} & 15.0{\tiny $\pm 1.1$}  & 63.0{\tiny $\pm 0.8$}  & 40.9{\tiny $\pm 0.9$} & 14.6{\tiny $\pm 1.5$}  & 43.9{\tiny $\pm 0.3$}  & 3.1{\tiny $\pm 0.9$} \\ 

          \multicolumn{1}{|l|}{\rule[-1.7ex]{0pt}{0ex}\textcolor{Blue}{SlotDiff.}}     & 76.3{\tiny $\pm 1.2$} & 89.3{\tiny $\pm 1.3$} & 94.1{\tiny $\pm 1.5$} & 63.5{\tiny $\pm 0.4$}  & 67.0{\tiny $\pm 1.7$}  & 80.2{\tiny $\pm 1.6$} &  6.6{\tiny $\pm 0.9$} & 56.3{\tiny $\pm 0.4$} & 39.5{\tiny $\pm 1.3$}& 12.0{\tiny $\pm 1.2$}  & 33.9{\tiny $\pm 2.2$} & 1.4{\tiny $\pm 0.4$} \\ 
       \cline{1-13}
      \end{tabular}}}
      \end{minipage}};
  \end{tikzpicture}
  \label{tab:ablation}
\end{figure}

\section{Analysing Representation Quality with Object-Centric Instance Retrieval}
\label{sec:pelim_analysis}

To evaluate how well existing methods capture object-centric detail, we compare \textcolor{DarkGoldenrod}{DINOv2} \citep{caron2021dino}, \textcolor{Purple}{DINOv3} \citep{simeoni2025dinov3}, \textcolor{DarkGreen}{CLIP} \citep{radford2021learning}, and \textcolor{Blue}{SlotDiffusion} \citep{wu2023slotdiffusion} representations in multi-object instance retrieval. 
We use the CLEVR and CLEVRTex datasets, which provide attribute labels (shape, size, material, colour), enabling us to probe whether these properties are faithfully preserved in the learned embeddings. 
Beyond single attributes, we also test combinations, since downstream applications often require correctly binding multiple attributes to the same object.  

Figure~\ref{fig:intro} illustrates the setup: the top row shows qualitative examples of top-4 retrievals for a single reference object, while the bottom row reports top-10 precision across individual attributes and attribute subsets. Formally, we compute means over $\mathcal{P}_i(\mathcal{X}) = \{A \in \mathcal{P}(\mathcal{X}): \lvert A \rvert = i\}$, where $\mathcal{X} = \{\text{shape, size, material, colour}\}$, and extend the evaluation to ordered permutations.  
We further present extended results in Table \ref{tab:ablation}.

\textbf{Colour is the weakest attribute across models.}  
\textcolor{DarkGoldenrod}{DINOv2} and \textcolor{Purple}{DINOv3} both perform strongly on shape (\textcolor{DarkGoldenrod}{95.4\%}, \textcolor{Purple}{96.9\%}) but fall sharply on colour (\textcolor{DarkGoldenrod}{40.8\%}, \textcolor{Purple}{44.5\%}). This indicates that newer self-supervised objectives have not fully addressed colour encoding. \textcolor{DarkGreen}{CLIP} improves colour substantially (\textcolor{DarkGreen}{66.8\%}), but at the cost of weaker shape and size performance compared to DINO models (\textcolor{DarkGreen}{83.8\%} shape, \textcolor{DarkGreen}{83.2\%} size). \textcolor{Blue}{SlotDiffusion} also retrieves colour better than \textcolor{DarkGoldenrod}{DINOv2} (\textcolor{Blue}{63.5\%} vs. \textcolor{DarkGoldenrod}{40.8\%}), yet underperforms on shape (\textcolor{Blue}{76.3\%}), suggesting a trade-off between geometric and surface-level features.  

\textbf{Slot-based methods struggle with attribute binding.}  
In multi-attribute retrieval, \textcolor{Blue}{SlotDiffusion} degrades rapidly: while colour can be matched in isolation, it fails to consistently bind multiple attributes to the same object. Visual inspection confirms that while \textcolor{DarkGoldenrod}{DINOv2} and \textcolor{Purple}{DINOv3} retrievals often fail only by colour mismatch, \textcolor{Blue}{SlotDiffusion} frequently retrieves objects that share little resemblance with the query, indicating difficulties in binding attributes.  

\textbf{CLIP is inconsistent across domains.}  
\textcolor{DarkGreen}{CLIP} shows relatively strong performance on single attributes like material (\textcolor{DarkGreen}{98.3\%}), but its precision collapses when tested on CLEVR multi-attribute settings, particularly when colour is included (\textcolor{DarkGreen}{14.6\%}). This suggests that \textcolor{DarkGreen}{CLIP}'s text-image alignment biases help capture some surface cues, but generalisation to synthetic multi-object scenes is weaker.  

Overall, these results reinforce the broader picture: self-supervised and slot-based representations excel at encoding geometry but consistently neglect surface-level cues such as colour and material, which are critical for disambiguating otherwise similar objects.
This motivates our method: instead of altering pre-training, which is expensive and dataset-specific, we introduce a complementary object-centric latent space.
We focus on \textcolor{DarkGoldenrod}{DINOv2}-S as the foundation, since we will show that it delivers the most faithful attribute binding overall and can be augmented most effectively.
By training a Variational Autoencoder (VAE) on image patches segmented from \textcolor{DarkGoldenrod}{DINOv2} features, and augmenting the global DINO embeddings with these latents, we obtain representations that recover fine-grained attributes while preserving global scene context.
\section{Enhancing Self-Supervised Representations with Object-Centric Latents}
\label{sec:method}
\FloatBarrier
\begin{figure}[htbp!]
    \centering
    \includegraphics[width=1.0\linewidth]{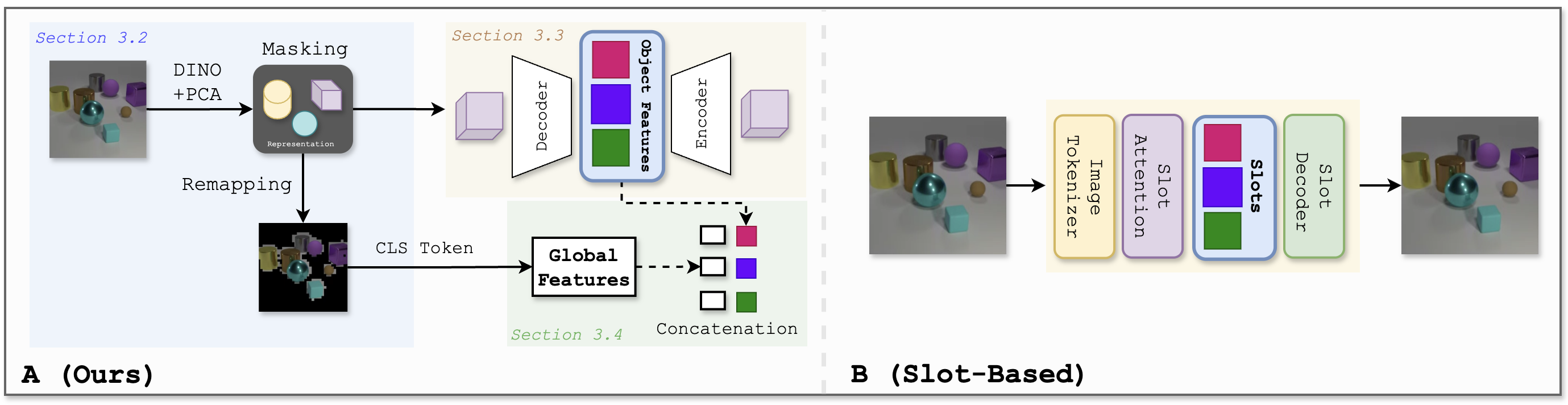}
    \caption{\textbf{Our method leverages the implicit general object understanding of DINOv2 to extract object-level features for multi-object instance retrieval.} 
    \textbf{B:} Traditional slot-based object-centric methods learn slot-representations via cross-attention which provide little inductive bias, while having to compress global and object-level features into a small set of latents. 
    \textbf{A:} 
    We propose combining general self-supervised features with learnt object-level features to obtain an improved latent representation. The object-level features are learnt from image patches 
    making training efficient and the latent space expressive.}
    \label{fig:method1}
\end{figure}
Our approach, illustrated in Figure~\ref{fig:method1}, enriches self-supervised features with object-level latents to better capture fine-grained attributes. We will show that DINOv2 features benefit from this the most. We proceed in four steps: (i) extract patch embeddings from DINOv2, (ii) segment objects using PCA, (iii) learn a latent space over object patches with a VAE, and (iv) combine global and local features into a joint representation.

\subsection{Preliminaries}
\label{sec:prelim}
Given an image $x \in \mathbb{R}^{h \times w \times c}$, we extract patch embeddings $y \in \mathbb{R}^{p \times n_y}$ from a pre-trained DINOv2 model, where $p$ is the number of patches. These embeddings form the basis for both segmentation and global features. To stabilise segmentation, we compute PCA over batches of $t=50$ images.

\subsection{Extracting Object-Level Features with PCA}
\label{sec:pca}

\begin{figure}[htbp!]
  \centering
  \includegraphics[scale=0.25]{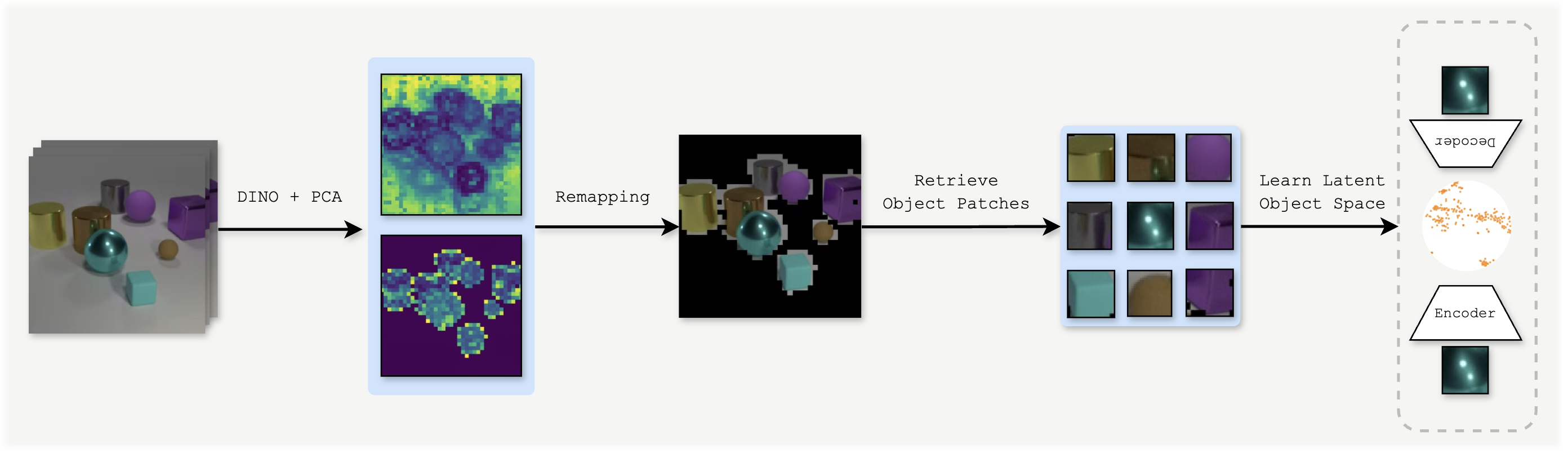}
  \caption{\textbf{Detailed view of extracting object-level features using PCA.} From $t$ images we create a segmentation mask which is used to extract the object image patches. We then 
  learn a latent space of the object patches with a VAE.}
  \label{fig:detailed_pca}
\end{figure}

We extract object-level patches from DINOv2 embeddings using a simple PCA-based segmentation procedure (Figure~\ref{fig:method1}). 
While DINOv3 can be used for this segmentation procedure as well, in practice we found DINOv2 to produce better segmentation for our use case.
This involves three steps: (i) separating foreground from background, (ii) refining object consistency, and (iii) remapping the mask to the original image to obtain patches. 
We describe the process in the following along with Figure \ref{fig:detailed_pca}.

\textbf{(i) Foreground–background separation.} 
We first flatten all patch embeddings $Y \in \mathbb{R}^{t \cdot p \times n_y}$ collected from a batch of $t$ images and apply PCA. The first principal component $Z' \in \mathbb{R}^{t \cdot p}$ reliably distinguishes background from foreground content, consistent with prior findings on DINO features \citep{simeoni2021localizing, wang2023cut}. A binary mask is obtained by thresholding against the median:
\begin{equation}
    \hat{M}^{\text{fg}} = \mathbbm{1}(Z' > \text{median}(Z')).
\end{equation}
This mask selects patches belonging to salient objects.

\textbf{(ii) Refining object consistency.} 
While this initial mask captures saliency, it can fragment objects. We therefore reapply PCA on the foreground embeddings $Y^{\text{fg}} = \hat{M}^{\text{fg}} \odot Y$, yielding a refined mask $M^{\text{fg}}$. This second pass improves spatial coherence, avoiding masks that cut across single objects.

\textbf{(iii) Remapping to image space.} 
Finally, we map the binary mask of patches back to the original image resolution. Each mask entry corresponds to a patch of size $s_h \times s_w$; we expand these back into image coordinates, producing a segmented image $x^{o}$ and a set of object patches
\begin{equation}
\rho = \{ \rho_1, \rho_2, \ldots, \rho_p \}, \quad \rho_i \in \mathbb{R}^{s_h \times s_w \times c}.
\end{equation}
These patches serve two purposes: (a) they provide a segmented image $x^o$ from which we later extract global features, and (b) they form the training input for the VAE in Section~\ref{sec:vae}.

\subsection{Learning Object-Level Latent Representations with a VAE}
\label{sec:vae}
To encode fine-grained object attributes, we train a Variational Autoencoder on the segmented patches. The encoder maps each patch $\rho$ to a latent vector $z \in \mathbb{R}^{n_z}$, while the decoder reconstructs the patch. The VAE loss is
\begin{equation}
\mathcal{L}_{\text{VAE}} = \mathbb{E}_{z \sim q_\phi(z|\rho)}[\log f_\theta(\rho|z)] 
- \beta D_{\text{KL}}(q_\phi(z|\rho) \,\|\, \mathcal{N}(0,1)),
\end{equation}
with a small $\beta$ to encourage attribute-specific clustering. This yields a set of object-level latents $z \in \mathbb{R}^{p \times n_z}$, which will be combined with SSL representations in the following step.

\subsection{Combining Global and Local Representations}
\label{sec:similarity}
We combine the global scene feature from the self-supervised model (CLS token in DINOs case) $\epsilon \in \mathbb{R}^{n_g}$ with the object-level latents. For each patch, we concatenate $\epsilon$ with its VAE latent to form
\begin{equation}
v = [\epsilon, z] \in \mathbb{R}^{p \times (n_g+n_z)}.
\end{equation}
This structured representation retains global context while injecting object-level detail. Retrieval is then performed by cosine similarity between $v$ and $v'$ from query and candidate images as 
shown in Appendix \ref{sec:similarities}

\section{Related Work: Object-Centric Learning and Instance Retrieval}
\paragraph{Slot-based object centric learning.}
Recent advances in object-centric representation learning (e.g., SlotAttention \citep{locatello2020slotattention}, MONet \citep{burgess2019monet}, SlotDiffusion \citep{wu2023slotdiffusion},  AdaSlot \citep{fan2024adaptive}, Slot-VAE \citep{wang2023slotvae}) 
have enabled models to decompose images into object-level features, making it possible to reason about individual objects within a scene.
These methods mostly focus on learning disentangled representations of individual objects by enforcing an object-level slot-based bottleneck via cross attention,
which then allows for faithful image reconstruction, generation of novel object configurations, or object segmentation. 
In order to extract the objects, the image is usually tokenised and an attention mechanism is applied to the tokens (e.g. SLATE \citep{singh2021illiterate}).
A slot decoder then reconstructs the original image from the slot representation.  STEVE \citep{singh2022simple} reconstructs the image from the slot-representation autoregressively to learn 
dependencies between different image patches i.e. tokens. 
More recent methods, have further exploited this by attempting to incorporate global scene level features into the representations \citep{chen2024gold} or by adding compositional reasoning into the slot-based representations \citep{jung2024learning}.


\paragraph{Combining self-supervised features with task-specific features.}
Recent methods such as SLATE \citep{singh2021illiterate} and STEVE \citep{singh2022simple} have leveraged vision transformers for object-centric learning to learn improved slot representations that also capture global scene information via attention mechanisms.
These methods have shown improved performance on object-centric tasks, but they require retraining the complete model, which can be computationally expensive and time-consuming. 
Similarly, DINOSAUR \citep{seitzer2023bridging} and GOLD \citep{chen2024gold} use DINO representations to provide better natural priors to learn object-centric representations on natural images. 
R$^2$Former \citep{zhu2023r2former} learns global correspondences between patches from a transformer pre-trained on ImageNet and then leverages local attention maps to rerank retrieved patches of candidate images. 
Zhang et al.\citep{zhang2023tale} combine learnt Stable Diffusion and DINOv2 features to enhance instance retrieval. We build on this work by combining representations that occupy different semantic spaces, unlike the previously studied Stable Diffusion and DINOv2 features which share similar characteristics. This integration of semantically distinct vectors (i.e. general features and object-level features) enhances our model's capabilities.

\paragraph{Assessing the quality of representations with instance retrieval.}
Instance retrieval is a challenging task that requires models to retrieve images with similar object attributes and configurations. Traditionally,
this task has been best solved by methods that leverage representations from pre-trained models, such as ResNet \citep{he2016deep} or vision transformers \citep{dosovitskiy2020image}.
Generally, instance retrieval depends on learning a good latent space or representation of the image, where retrievals are ranked on a similarity measure \citep{chen2021instance}. 
Naturally, instance retrieval has been used to evaluate the quality of representations in self-supervised models \citep{caron2021dino, oquab2023DINOv2}. Since we are interested in evaluating the quality
of the representations for multi-object understanding, we choose the task of multi-object instance retrieval.
Beyond traditional vision applications, multi-object instance retrieval plays a crucial role in goal-conditioned reinforcement learning (GCRL) and robotic tasks \citep{zhen2024survey}. 
In GCRL, agents often rely on goal images to specify desired configurations of objects in the environment, 
particularly in tasks requiring manipulation or navigation in multi-object settings \citep{paine2019making}. 
Similar challenges are present in hierarchical reinforcement learning (HRL), where sub-goals often involve interacting with specific object sets. 
Robust multi-object retrieval can inform option selection and improve learning efficiency \citep{hafner2022director}. Moreover, the emergence
of embodied agents has seen the need for pre-trained representations
that understand nuance in their environment \citep{mazoure2023embodied,szot2024multimodal}.


\section{Experiments}
\subsection{Datasets}

A key challenge in evaluating object-centric representations is finding datasets that expose systematic variation in object attributes while remaining relevant for real-world tasks. 
Most natural-image benchmarks lack consistent multi-object scenes with fine-grained labels, while purely synthetic datasets may be dismissed as too controlled. 
We therefore combine synthetic datasets for diagnostic evaluation with a real-world dataset for external validation.  

\textbf{CLEVR} \citep{johnson2017clevr} provides rendered 3D scenes where objects vary along shape, size, material, and colour. 
It offers cleanly disentangled factors that make it possible to probe which attributes are preserved in the embeddings.  

\textbf{CLEVRTex} \citep{karazija2021clevrtex} extends CLEVR by introducing realistic textures, increasing visual complexity and making attribute retrieval more challenging. 
Together, CLEVR and CLEVRTex serve as controlled testbeds to diagnose representation quality in multi-object settings.  

\textbf{Stanford Cars} \citep{krause2013stanford} complements these synthetic datasets with a real-world benchmark. 
Many car classes share near-identical geometry and differ only in colour and surface-level details, which directly tests whether representations capture the appearance-level attributes needed to disambiguate objects.  

For training, we fit the VAE on $2{,}000$ CLEVR/CLEVRTex images segmented into object patches. 
Evaluation is performed on $500$ query images and $5{,}000$ candidates for retrieval, with Stanford Cars used only for evaluation under its standard splits.

\subsection{Experimental Setup}

\textbf{Retrieval tasks.}  
We evaluate representations on a retrieval task where the goal is to recover objects with matching attributes given a reference image. 
For CLEVR and CLEVRTex, attribute labels allow systematic evaluation of \textbf{shape}, \textbf{size}, \textbf{material}, and \textbf{colour} (CLEVR) or \textbf{shape}, \textbf{size}, and \textbf{material} (CLEVRTex, with material combining texture and colour). 
For Stanford Cars, which lacks attribute annotations, we focus on appearance-level cues by computing the average deviation in RGB pixel values to assess colour consistency. 
Dataset splits are reported in Appendix~\ref{sec:additional}, Table~\ref{tab:splits}.  

\textbf{Part I: Synthetic diagnostic benchmarks.}  
On CLEVR and CLEVRTex, we carry out both (i) \emph{performance evaluation} and (ii) \emph{attribute-level analysis}.  
First, we measure retrieval quality using Top-10 precision, weighted precision, and error rate across all attribute subsets, probing how well each model preserves object identity.  
Second, we perform an ablation study which was presented in Table \ref{tab:ablation} to examine retrieval fidelity for individual attributes and combinations, highlighting failures in binding surface-level cues (e.g.\ colour, texture).  
In this context, we analyse the effect of using standalone VAE features and show whether VAE-augmented variants of DINOv2, DINOv3, CLIP, and SlotDiffusion improve attribute alignment.  
Finally, we test whether increasing backbone capacity (DINOv2-L) helps encode surface-level features.  
 
\textbf{Part II: Real-world benchmark.}  
To validate beyond synthetic settings, we repeat our retrieval experiments on the Stanford Cars dataset.  
Here, cars of the same class often share nearly identical geometry but differ in colour and surface details, providing a natural testbed for appearance-sensitive retrieval.  
This complements the controlled synthetic benchmarks and demonstrates that the limitations we identify are not confined to toy datasets.  
In particular, we measure the \emph{average Euclidean distance in colour space} between the query and retrieved instances.

\textbf{Evaluation metrics.}  
We report three metrics:  

(i) \emph{Top-10 Precision: }fraction of the first ten retrieved images that match the reference attributes;  

(ii) \emph{Weighted Precision: } same as above, but weighted by retrieval rank ($1/i$ for rank $i$), emphasising higher-ranked matches;  

(iii) \emph{Error Rate:} percentage of queries for which no correct retrieval is found in the top ten (for CLEVRTex, normalised for queries with fewer than ten valid candidates).  

\textbf{Baselines.}  
For each method we run seven trials with 50 query images and 5,000 candidate images, sampling queries without replacement. SlotDiffusion is pre-trained separately on CLEVR and CLEVRTex.
Seeds are chosen at random and kept consistent across models. We report standard deviation across runs.

\subsection{CLEVR and CLEVRTex: Multi-Object Instance Retrieval} 
\begin{figure}[htbp!]
  
  \captionsetup{name=Table, type=table}
  \caption{\textbf{Oh-A-DINOv2 improves over SSL and slot-based features for multi-object instance retrieval.} 
  Our method outperforms both self-supervised features and slot-based approaches. The table shows that Oh-A-DINOv2 achieves higher precision and lower error rates on CLEVR and CLEVRTex, indicating that combining DINOv2 with object-level VAE features addresses the limitations of pure SSL features and slot-attention methods.}
  
    \centering
    \begin{tikzpicture}
        
        


        \node at (0,3) {

        \begin{minipage}{1.0\textwidth}            
            {\fontfamily{pcr}\selectfont \footnotesize
            \scalebox{0.653}{
            \setlength{\arrayrulewidth}{0.7pt}  
            \setlength{\tabcolsep}{3pt}  

            \hspace*{2.5cm}
            \begin{tabular}{c|ccc|ccc|}
            \cline{2-7}
            & \multicolumn{3}{c|}{\rule[-1.2ex]{0pt}{4.1ex}CLEVR} & \multicolumn{3}{c|}{CLEVRTex} \\ \cline{2-7}
            \multicolumn{1}{c|} {\textbf{Main Results (\%)}}  & {\rule[-1.7ex]{0pt}{4.6ex}\scriptsize\makecell{\emph{Top-10} \\ \emph{Precision}$\uparrow$}} & {\scriptsize\makecell{\emph{Weighted} \\ \emph{Precision}$\uparrow$}} & {\scriptsize\makecell{\emph{Error} \\ \emph{Rate}$\downarrow$}} &  {\scriptsize\makecell{\emph{Top-10} \\ \emph{Precision}$\uparrow$}} & {\scriptsize\makecell{\emph{Weighted} \\ \emph{Precision}$\uparrow$}} & {\scriptsize\makecell{\emph{Error} \\ \emph{Rate}$\downarrow$}}  \\ \cline{1-7}
            \multicolumn{1}{|l|}{\rule[-1.7ex]{0pt}{4.6ex}\textcolor{DarkRed}{Oh-A-DINOv2\textbf{(Ours)}}} & \textbf{56.4{\tiny $\pm 2.0$}} & \textbf{49.0{\tiny $\pm 2.2$}} & \textbf{1.0{\tiny $\pm 0.0$}} & \textbf{20.3{\tiny $\pm 2.6$}} & \textbf{11.2{\tiny $\pm 1.4$}}  & \textbf{41.0{\tiny $\pm 1.0$}}  \\ 
            \multicolumn{1}{|l|}{\rule[-1.7ex]{0pt}{0ex}\textcolor{DarkGoldenrod}{DINOv2}}     & 13.0{\tiny $\pm 1.1$} & 13.8{\tiny $\pm 1.2$} & 28.0{\tiny $\pm 1.0$}  & 12.2{\tiny $\pm 2.1$} & 8.0{\tiny $\pm 1.3$} & 55.6{\tiny $\pm 0.7$}  \\ 
            \multicolumn{1}{|l|}{\rule[-1.7ex]{0pt}{0ex}\textcolor{Purple}{DINOv3}}     & 13.7{\tiny $\pm 1.2$} & 14.5{\tiny $\pm 1.3$} & 28.0{\tiny $\pm 2.0$}  & 4.2{\tiny $\pm 1.1$} & 1.5{\tiny $\pm 0.3$} & 82.4{\tiny $\pm 1.9$}  \\ 
            \multicolumn{1}{|l|}{\rule[-1.7ex]{0pt}{0ex}\textcolor{DarkGreen}{CLIP}}     & 14.7{\tiny $\pm 1.5$} & 16.3{\tiny $\pm 1.6$} & 29.2{\tiny $\pm 1.9$}  & 2.9{\tiny $\pm 1.2$} & 0.7{\tiny $\pm 0.2$} & 90.6{\tiny $\pm 0.4$}  \\ 
            \multicolumn{1}{|l|}{\rule[-1.7ex]{0pt}{0ex}\textcolor{Blue}{SlotDiff.}}     & 12.0{\tiny $\pm 1.2$} & 12.7{\tiny $\pm 1.3$} & 33.6{\tiny $\pm 1.5$} & 1.4{\tiny $\pm 0.4$}  & 0.9{\tiny $\pm 0.2$}  & 93.4{\tiny $\pm 0.2$} \\ 
            \cline{1-7}
        \end{tabular}}}       
        \end{minipage}};
    \end{tikzpicture}
    \label{tab:retrieval}
\end{figure}
\begin{figure}[htbp!]
  \centering
  \includegraphics[width=0.85\linewidth]{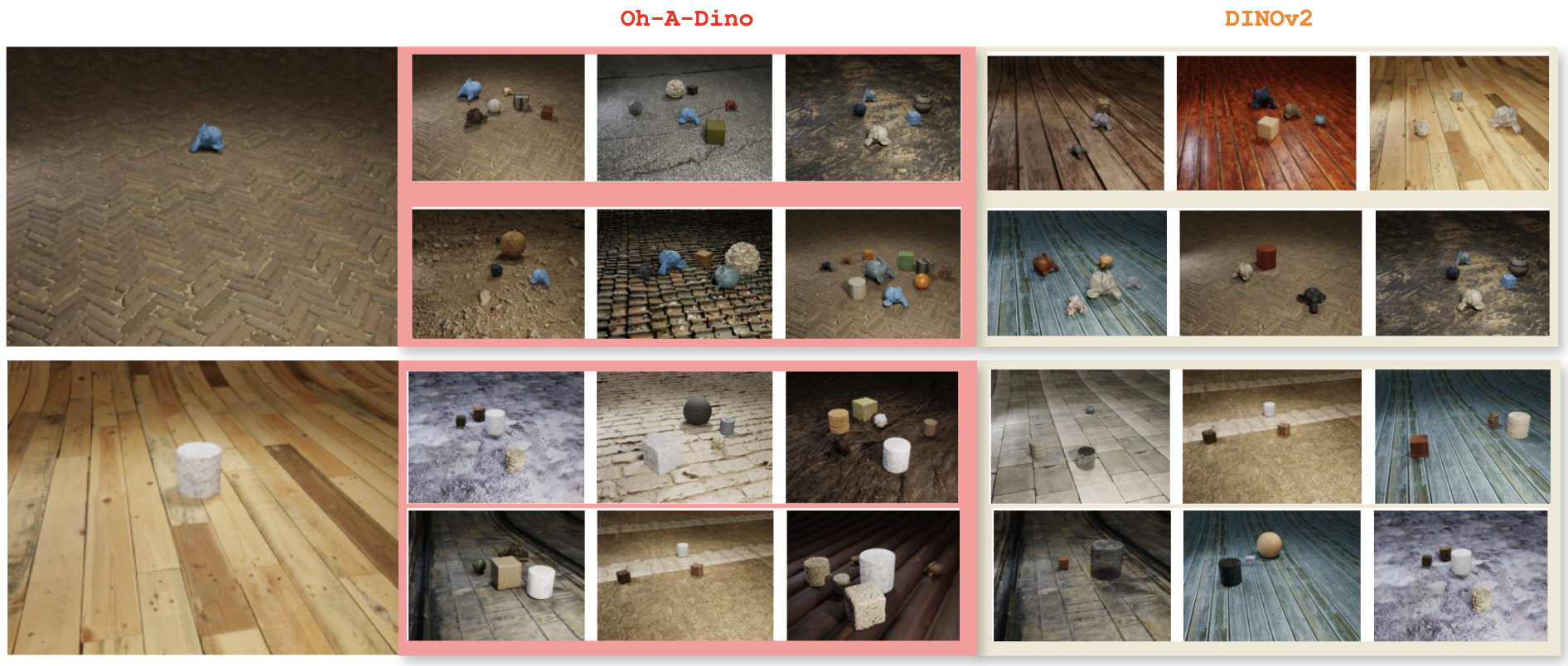}
  \caption{\textbf{Visualisations of retrievals for CLEVRTex, augmenting with VAE object level features allows for retrieval in complex multi-object scenes.} 
  \textcolor{DarkRed}{Oh-Ah-Dino} consistently retrieves the object in question with the correct properties such as the shape, size and material. 
  \textcolor{DarkGoldenrod}{DINOv2} on the other hand often retrieves the correct object but with the wrong texture. Further visualisations, also for other methods
  can be found in Appendix \ref{sec:vis}.} 
  \label{fig:clevrtex_vis}
\end{figure}

\textbf{Our approach improves multi-object instance retrieval.} 
We show in Table \ref{tab:retrieval} that the learnt object features are expressive and complement the DINOv2 representations well. 
On CLEVR, \textcolor{DarkRed}{Oh-A-DINOv2} achieves a Top-10 Precision of \textbf{56.4\%}, compared to only \textcolor{DarkGoldenrod}{13.0\%} for \textcolor{DarkGoldenrod}{DINOv2}, \textcolor{Purple}{13.7\%} for \textcolor{Purple}{DINOv3}, \textcolor{DarkGreen}{14.7\%} for \textcolor{DarkGreen}{CLIP}, and \textcolor{Blue}{12.0\%} for \textcolor{Blue}{SlotDiffusion}. 
On CLEVRTex, \textcolor{DarkRed}{Oh-A-DINOv2} reaches \textbf{20.3\%}, substantially higher than the \textcolor{DarkGoldenrod}{12.2\%} of \textcolor{DarkGoldenrod}{DINOv2}, the \textcolor{Purple}{4.2\%} of \textcolor{Purple}{DINOv3}, the \textcolor{DarkGreen}{2.9\%} of \textcolor{DarkGreen}{CLIP}, and the \textcolor{Blue}{1.4\%} of \textcolor{Blue}{SlotDiffusion}. 
Interestingly, \textcolor{Purple}{DINOv3} performs markedly worse than \textcolor{DarkGoldenrod}{DINOv2} on CLEVRTex (4.2\% vs.\ 12.2\%), suggesting that the newer model is less robust to textured multi-object scenes. 
These results together with Table \ref{tab:ablation} confirm that SSL features (DINO/CLIP) fail to capture non-geometric attributes, slot-based features collapse under increased scene complexity, and that combining SSL (specifically DINOv2) with object-level VAE representations directly addresses both shortcomings.

\subsubsection{The Role of the VAE features}
\begin{figure}[htbp!]
  \centering
  \captionsetup{name=Table, type=table}
  \caption{\textbf{Object-level VAE features enhance color and multi-attribute retrieval.} 
While the standalone \textcolor{magenta}{VAE} improves fine-grained color retrieval on both CLEVR and CLEVRTex, only \textcolor{DarkRed}{Oh-A-DINOv2} effectively integrates this signal with DINOv2 scene-level features. 
This leads to substantial gains in multi-attribute retrieval, where pure VAE features or SSL features (\textcolor{DarkGoldenrod}{DINOv2}) alone fall short.}
  \begin{tikzpicture}
      
      
      


      \node at (-7.2,4) {

      \begin{minipage}{1.0\textwidth}            
          {\fontfamily{pcr}\selectfont \footnotesize
          \scalebox{0.653}{
          \setlength{\arrayrulewidth}{0.7pt}  
          \setlength{\tabcolsep}{3pt}  

          \hspace*{2.7cm}
          \begin{tabular}{c|ccc|ccc|}
          \cline{2-7}
          & \multicolumn{3}{c|}{\rule[-1.2ex]{0pt}{4.1ex}CLEVR} & \multicolumn{3}{c|}{CLEVRTex}\\ \cline{2-7}
          \multicolumn{1}{c|} {\textbf{VAE Features (\%)}}  & {\rule[-1.7ex]{0pt}{4.6ex}\scriptsize\makecell{\emph{Top-10} \\ \emph{Precision}$\uparrow$}} & {\scriptsize\makecell{\emph{Weighted} \\ \emph{Precision}$\uparrow$}} & {\scriptsize\makecell{\emph{Error} \\ \emph{Rate}$\downarrow$}} &  {\scriptsize\makecell{\emph{Top-10} \\ \emph{Precision}$\uparrow$}} & {\scriptsize\makecell{\emph{Weighted} \\ \emph{Precision}$\uparrow$}} & {\scriptsize\makecell{\emph{Error} \\ \emph{Rate}$\downarrow$}} \\ \cline{1-7}
          \multicolumn{1}{|l|}{\rule[-1.7ex]{0pt}{4.6ex}\textcolor{magenta}{VAE\textbf{(Ours)}}}     & 53.1{\tiny $\pm 3.6$} & 29.2{\tiny $\pm 2.7$} & 2.0{\tiny $\pm 1.1$}  & 11.7{\tiny $\pm 3.7$} & 4.1{\tiny $\pm 1.3$} & 60.0{\tiny $\pm 3.4$}  \\ 
          \multicolumn{1}{|l|}{\rule[-1.7ex]{0pt}{0ex}\textcolor{DarkRed}{Oh-A-DINOv2\textbf{(Ours)}}} & \textbf{56.4{\tiny $\pm 2.0$}} & \textbf{49.0{\tiny $\pm 2.2$}} & \textbf{1.0{\tiny $\pm 0.0$}} & \textbf{20.3{\tiny $\pm 2.6$}} & \textbf{11.2{\tiny $\pm 1.4$}}  & \textbf{41.0{\tiny $\pm 1.0$}}\\ 
          \multicolumn{1}{|l|}{\rule[-1.7ex]{0pt}{0ex}\textcolor{DarkGoldenrod}{DINOv2}}     & 13.0{\tiny $\pm 1.1$} & 13.8{\tiny $\pm 1.2$} & 28.0{\tiny $\pm 1.0$}  & 12.2{\tiny $\pm 2.1$} & 8.0{\tiny $\pm 1.3$} & 55.6{\tiny $\pm 0.7$}  \\ 

          \cline{1-7}
      \end{tabular}}}       
      \end{minipage}};
  \end{tikzpicture}

  \begin{tikzpicture}

      \node[align=left] (bg) at  (-2.25,1.85) {{\fontfamily{pcr}\selectfont\tiny Multi-Attribute}};
      \node[align=left] (bg) at  (-8.5,1.85) {{\fontfamily{pcr}\selectfont\tiny Single-Attribute}};
      
      
      


      \node at (-7.2,3) {

      \begin{minipage}{1.0\textwidth}            
          {\fontfamily{pcr}\selectfont \footnotesize
          \scalebox{0.6}{
          \setlength{\arrayrulewidth}{0.7pt}  
          \setlength{\tabcolsep}{3pt}  

          \hspace*{0.3cm}
          \begin{tabular}{c|cccc|ccc!{\vrule width 2pt}ccc|cc|}
          \cline{2-13}
          & \multicolumn{4}{c|}{\rule[-1.2ex]{0pt}{4.1ex}CLEVR} & \multicolumn{3}{c!{\vrule width 2pt}}{CLEVRTex} & \multicolumn{3}{c|}{CLEVR} & \multicolumn{2}{c|}{CLEVRTex} \\ \cline{2-13}
          \multicolumn{1}{c|} {\makecell{\textbf{Attrib.\;Ablation} \\ \tiny{Top-10 Precision (\%)}}}  & \rule[-1.7ex]{0pt}{4.6ex} \emph{Shape} & \emph{Size} & \emph{Mat.}& \emph{Col.} & \emph{Shape} & \emph{Size} & \emph{Mat.}&  \emph{$\mathcal{P}_2(\mathcal{X})$} & \emph{$\mathcal{P}_3(\mathcal{X})$} & \emph{$\mathcal{P}_3(\mathcal{X})$+$C$} & \emph{$\mathcal{P}_2(\mathcal{X})$} & \emph{$\mathcal{P}_2(\mathcal{X})$+$C$} \\ \cline{1-13}
          \multicolumn{1}{|l|}{\rule[-1.7ex]{0pt}{4.6ex}\textcolor{magenta}{VAE\textbf{(Ours)}}}     & 93.4{\tiny $\pm 1.2$} & \textbf{99.2}{\tiny $\pm 0.4$} & \textbf{99.6}{\tiny $\pm 0.3$}  & \textbf{98.7}{\tiny $\pm 0.4$} & 89.0{\tiny $\pm 0.9$} & \textbf{94.4}{\tiny $\pm 0.8$} & 43.5{\tiny $\pm 5.3$}  & 82.8{\tiny $\pm 0.5$}  & 63.3{\tiny $\pm 0.5$} & 53.1{\tiny $\pm 3.6$} &  52.5{\tiny $\pm 0.1$}  & 11.7{\tiny $\pm 3.7$} \\ 
          \multicolumn{1}{|l|}{\rule[-1.7ex]{0pt}{0ex}\textcolor{DarkRed}{Oh-A-DINOv2\textbf{(Ours)}}} & 94.7{\tiny $\pm 2.0$} & 96.7{\tiny $\pm 2.2$} & 98.6{\tiny $\pm 0.0$} & 96.6{\tiny $\pm 2.6$} & \textbf{90.7}{\tiny $\pm 2.1$}  & 85.3{\tiny $\pm 1.5$} & \textbf{45.8{\tiny $\pm 3.9$}}  & \textbf{85.2{\tiny $\pm 2.1$}} & \textbf{72.8{\tiny $\pm 1.4$}}& \textbf{56.4{\tiny $\pm 1.6$}}  & \textbf{56.0}{\tiny $\pm 0.2$} & \textbf{20.3}{\tiny $\pm 2.6$} \\ 

          \multicolumn{1}{|l|}{\rule[-1.7ex]{0pt}{0ex}\textcolor{DarkGoldenrod}{DINOv2}}     & \textbf{95.4}{\tiny $\pm 1.1$} & 81.2{\tiny $\pm 1.2$} & 96.1{\tiny $\pm 1.0$}  & 40.8{\tiny $\pm 2.1$} & 87.3{\tiny $\pm 3.8$} & 80.3{\tiny $\pm 2.6$} & 24.2{\tiny $\pm 1.9$}  & 76.2{\tiny $\pm 3.1$}  & 62.8{\tiny $\pm 1.2$} & 13.0{\tiny $\pm 1.1$}  & 54.0{\tiny $\pm 0.1$}  & 12.2{\tiny $\pm 2.1$} \\ \cline{1-13}
      \end{tabular}}}
      \end{minipage}};
  \end{tikzpicture}
  \label{tab:vae_retrieval}
\end{figure}

\textbf{Object-level features improve multi-object instance retrieval when combined with DINOv2.}
Table \ref{tab:vae_retrieval} (bottom) shows that individual attribute prediction improves significantly with the learned object-level representations \textcolor{magenta}{VAE} for both CLEVR and CLEVRTex.
This improves color prediction in \textcolor{DarkRed}{Oh-A-DINO}.
However, Table \ref{tab:vae_retrieval} (top) reveals that only the Oh-A-DINO combination manages to utilise this color signal to improve multi-attribute retrieval.
When comparing \textcolor{DarkRed}{Oh-A-DINO} and \textcolor{magenta}{VAE} in CLEVR, even with a smaller performance gap in CLEVR Top-$10$ precision, our method delivers substantially better weighted precision, indicating that DINO's scene features help retrieve more consistent objects.  
This is further validated in Appendix \ref{sec:vis}, where we show that Oh-A-DINO consistently performs the closest retrievals regardless of exact matching.

\begin{figure}[htbp!]
  \centering
  \captionsetup{name=Table, type=table}
  \caption{\textbf{Ablation on VAE augmentation and model capacity.} 
  (left) Augmenting SSL and slot-based features with object-level VAE latents improves retrieval across methods, 
  with the largest gains observed for \textcolor{DarkRed}{Oh-A-DINOv2}. 
  (right) Scaling backbone capacity from \textcolor{DarkGoldenrod}{DINOv2-S} to \textcolor{DarkGoldenrod}{DINOv2-L} provides only minor improvements.}
  \begin{minipage}{0.48\textwidth}
    {\fontfamily{pcr}\selectfont \footnotesize
              \scalebox{0.503}{
              \setlength{\arrayrulewidth}{0.7pt}  
              \setlength{\tabcolsep}{3pt}  
    
              \hspace*{-0.8cm}
              \begin{tabular}{c|ccc|ccc|}
              \cline{2-7}
              & \multicolumn{3}{c|}{\rule[-1.2ex]{0pt}{4.1ex}CLEVR} & \multicolumn{3}{c|}{CLEVRTex} \\ \cline{2-7}
              \multicolumn{1}{c|} {\textbf{\makecell{Augmenting \\ with VAE (\%)}}}  & {\rule[-1.7ex]{0pt}{4.6ex}\scriptsize\makecell{\emph{Top-10} \\ \emph{Precision}$\uparrow$}} & {\scriptsize\makecell{\emph{Weighted} \\ \emph{Precision}$\uparrow$}} & {\scriptsize\makecell{\emph{Error} \\ \emph{Rate}$\downarrow$}} &  {\scriptsize\makecell{\emph{Top-10} \\ \emph{Precision}$\uparrow$}} & {\scriptsize\makecell{\emph{Weighted} \\ \emph{Precision}$\uparrow$}} & {\scriptsize\makecell{\emph{Error} \\ \emph{Rate}$\downarrow$}} \\ \cline{1-7}
              \multicolumn{1}{|l|}{\rule[-1.7ex]{0pt}{4.6ex}\textcolor{DarkRed}{Oh-A-DINOv2\textbf{(Ours)}}} & \textbf{56.4{\tiny $\pm 2.0$}} & \textbf{49.0{\tiny $\pm 2.2$}} & \textbf{1.0{\tiny $\pm 0.0$}} & \textbf{20.3{\tiny $\pm 2.6$}} & \textbf{11.2{\tiny $\pm 1.4$}}  & \textbf{41.0{\tiny $\pm 1.0$}}  \\ 
              \multicolumn{1}{|l|}{\rule[-1.7ex]{0pt}{0ex}\textcolor{Purple}{DINOv3}-\textcolor{magenta}{VAE}} & 34.8{\tiny $\pm 2.6$} & 25.2{\tiny $\pm 1.9$} & 10.8{\tiny $\pm 1.8$} & 5.7{\tiny $\pm 1.6$} & 1.0{\tiny $\pm 0.2$}  & 82.0{\tiny $\pm 1.8$} \\ 
              \multicolumn{1}{|l|}{\rule[-1.7ex]{0pt}{0ex}\textcolor{DarkGreen}{CLIP}-\textcolor{magenta}{VAE}} & 55.0{\tiny $\pm 2.4$} & 31.8{\tiny $\pm 1.5$} & 1.0{\tiny $\pm 0.0$} & 3.1{\tiny $\pm 1.0$} & 0.9{\tiny $\pm 0.2$} & 86.4{\tiny $\pm 1.2$} \\ 
              \multicolumn{1}{|l|}{\rule[-1.7ex]{0pt}{0ex}\textcolor{Blue}{SlotDiff.}-\textcolor{magenta}{VAE}}     & 14.3{\tiny $\pm 1.5$} & 6.8{\tiny $\pm 0.6$} & 30.6{\tiny $\pm 2.0$} & 2.8{\tiny $\pm 0.1$}  & 0.8{\tiny $\pm 0.0$}  & 87.2{\tiny $\pm 1.7$}  \\ 
              \cline{1-7}
          \end{tabular}}}               
  \end{minipage}
  \hfill
  \begin{minipage}{0.48\textwidth}
    {\fontfamily{pcr}\selectfont \footnotesize
              \scalebox{0.503}{
              \setlength{\arrayrulewidth}{0.7pt}  
              \setlength{\tabcolsep}{3pt}  
    
              \hspace*{-0.8cm}
              \begin{tabular}{c|ccc|ccc|}
              \cline{2-7}
              & \multicolumn{3}{c|}{\rule[-1.2ex]{0pt}{4.1ex}CLEVR} & \multicolumn{3}{c|}{CLEVRTex} \\ \cline{2-7}
              \multicolumn{1}{c|} {\textbf{Model Capacity (\%)}}  & {\rule[-1.7ex]{0pt}{4.6ex}\scriptsize\makecell{\emph{Top-10} \\ \emph{Precision}$\uparrow$}} & {\scriptsize\makecell{\emph{Weighted} \\ \emph{Precision}$\uparrow$}} & {\scriptsize\makecell{\emph{Error} \\ \emph{Rate}$\downarrow$}} &  {\scriptsize\makecell{\emph{Top-10} \\ \emph{Precision}$\uparrow$}} & {\scriptsize\makecell{\emph{Weighted} \\ \emph{Precision}$\uparrow$}} & {\scriptsize\makecell{\emph{Error} \\ \emph{Rate}$\downarrow$}} \\ \cline{1-7}
              \multicolumn{1}{|l|}{\rule[-1.7ex]{0pt}{4.6ex}\textcolor{DarkRed}{Oh-A-DINOv2\textbf{(Ours)}}} & \textbf{56.4{\tiny $\pm 2.0$}} & \textbf{49.0{\tiny $\pm 2.2$}} & \textbf{1.0{\tiny $\pm 0.0$}} & \textbf{20.3{\tiny $\pm 2.6$}} & \textbf{11.2{\tiny $\pm 1.4$}}  & \textbf{41.0{\tiny $\pm 1.0$}}  \\ 
              \multicolumn{1}{|l|}{\rule[-1.7ex]{0pt}{0ex}\textcolor{DarkGoldenrod}{DINOv2-S}}     & 13.0{\tiny $\pm 1.1$} & 13.8{\tiny $\pm 1.2$} & 28.0{\tiny $\pm 1.0$}  & 12.2{\tiny $\pm 2.1$} & 8.0{\tiny $\pm 1.3$} & 55.6{\tiny $\pm 0.7$}  \\ 
              \multicolumn{1}{|l|}{\rule[-1.7ex]{0pt}{0ex}\textcolor{DarkGoldenrod}{DINOv2-L}}     & 13.3{\tiny $\pm 1.2$} & 14.8{\tiny $\pm 1.1$} & 24.0{\tiny $\pm 2.0$}  & 13.2{\tiny $\pm 1.8$} & 9.0{\tiny $\pm 1.1$} & 50.6{\tiny $\pm 0.9$}  \\ 
              \cline{1-7}
          \end{tabular}}}       
  \end{minipage}
  \label{tab:ablations}
  \end{figure}
  
\subsubsection{Augmenting other SSL models with VAE latent features}
\textbf{Augmenting SSL features with VAE latents is effective, but depends on the backbone.} 
Table~\ref{tab:ablations} (left) further compares augmentations of different SSL and slot-based features with our learnt VAE latents. 
For \textcolor{Purple}{DINOv3}-\textcolor{magenta}{VAE}, improvements on CLEVR are moderate (34.8\% Top-10 Precision) and largely vanish on CLEVRTex. 
In contrast, \textcolor{DarkGreen}{CLIP}-\textcolor{magenta}{VAE} shows strong gains on CLEVR (55.0\% Top-10 Precision, comparable to \textcolor{DarkRed}{Oh-A-DINOv2}) but only minor improvements on CLEVRTex. 
Finally, combining \textcolor{Blue}{SlotDiffusion} with either \textcolor{DarkGoldenrod}{DINOv2} or \textcolor{magenta}{VAE} provides negligible benefit, suggesting that slot-based representations are not sufficiently disentangled to compose well with object-level latents. 
Overall, these results highlight that while VAE augmentation consistently helps SSL features, the degree of improvement depends strongly on the underlying representation, with \textcolor{DarkRed}{DINOv2} benefiting the most.

\subsubsection{Effect of model capacity}
\textbf{Scaling up SSL model capacity does not resolve the retrieval bottleneck.} 
Table~\ref{tab:ablations}(right) compares \textcolor{DarkGoldenrod}{DINOv2-S} with its larger variant \textcolor{DarkGoldenrod}{DINOv2-L}. 
While \textcolor{DarkGoldenrod}{DINOv2-L} achieves a slight improvement over \textcolor{DarkGoldenrod}{DINOv2-S} on both CLEVR (Top-10 Precision $13.3\%$ vs.\ $13.0\%$) and CLEVRTex ($13.2\%$ vs.\ $12.2\%$), the gains are marginal and far from closing the gap to \textcolor{DarkRed}{Oh-A-DINOv2} (\textbf{56.4\%} on CLEVR, \textbf{20.3\%} on CLEVRTex). 
This suggests that simply increasing model capacity within SSL frameworks does not lead to more disentangled or attribute-sensitive object representations. 
By contrast, augmenting with object-level VAE features directly addresses the compositional limitations of SSL features, providing consistent improvements irrespective of backbone size.

\section{Stanford Cars: Real-World Instance Retrieval}
\begin{figure*}[htbp!]
  \centering
  \begin{subfigure}[t]{0.12\textwidth}
    \centering
    \begin{tikzpicture}
      \node at (0,0) {
        \raisebox{1.3cm}{
      \begin{minipage}{1.0\textwidth}            
        {\fontfamily{pcr}\selectfont \footnotesize
        \scalebox{0.6}{ 
        \setlength{\arrayrulewidth}{0.7pt}  
        \setlength{\tabcolsep}{3pt}         
        \renewcommand{\arraystretch}{1.2}   
        \hspace*{-0.9cm}
        \begin{tabular}{c|c|}
          \cline{2-2}
          & \multicolumn{1}{c|}{\rule[-1.2ex]{0pt}{4.1ex}Stanford Cars} \\ \cline{2-2}
          \multicolumn{1}{c|}{\textbf{Main Results (\%)}}  
          & {\scriptsize\makecell{\rule[-1.7ex]{0pt}{4.6ex}\emph{Distance} \\ \emph{to Ref Colour}$\downarrow$}} \\ \cline{1-2}
          \multicolumn{1}{|l|}{\rule[-1.7ex]{0pt}{4.6ex}\textcolor{DarkRed}{Oh-A-DINOv2\textbf{(Ours)}}} & \textbf{0.432{\tiny $\pm 0.011$}} \\ 
          \multicolumn{1}{|l|}{\rule[-1.7ex]{0pt}{0ex}\textcolor{DarkGoldenrod}{DINOv2}} & 0.512{\tiny $\pm 0.014$} \\ 
          \multicolumn{1}{|l|}{\rule[-1.7ex]{0pt}{0ex}\textcolor{DarkGreen}{CLIP}} & 0.524{\tiny $\pm 0.021$} \\ 
          \cline{1-2}
        \end{tabular}}}     
      \end{minipage}}};
    \end{tikzpicture}
  \end{subfigure}
  \hfill
  \begin{subfigure}[t]{0.7\textwidth}
    \centering
    \hspace*{0cm}
    \includegraphics[width=\linewidth]{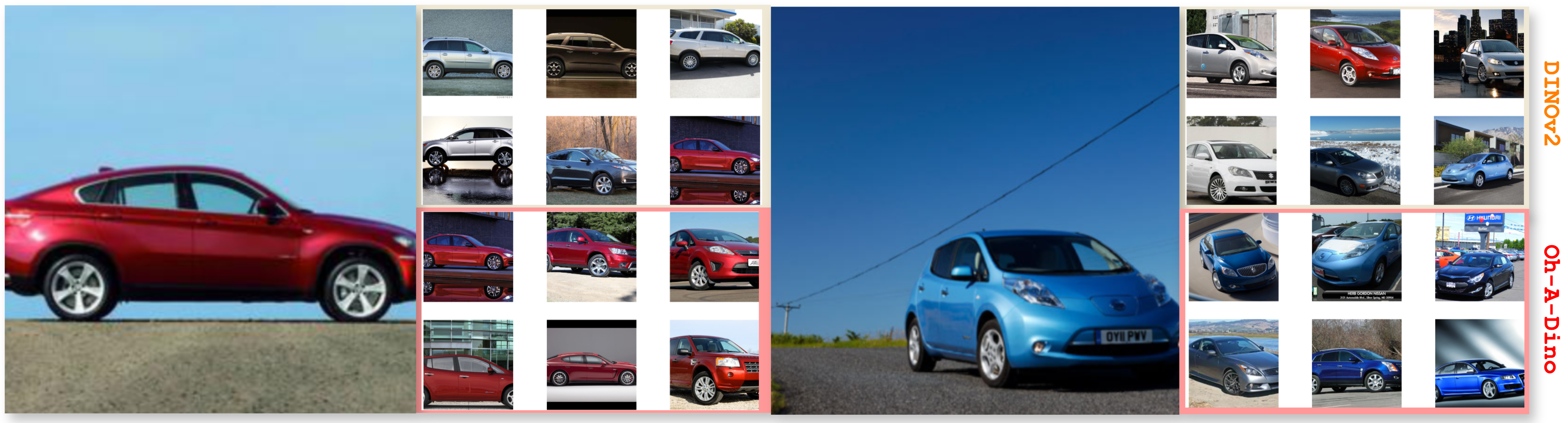}
  \end{subfigure}

  \caption{\textbf{Our method improves over baselines and leverages DINOv2’s general object understanding.} 
  (left) Table: comparison against baselines on Stanford Cars shows that our method achieves lower distance to the colours of the reference image. 
  (right) Oh-A-Dino can consistently retrieve cars which match orientation and shape while matching colour and material opposed to DINOv2.}
  \label{fig:cars_combined}
\end{figure*}
\textbf{VAE features help retrieve surface level attributes more faithfully for Stanford Cars.} 
Figure~\ref{fig:cars_combined} (left) shows that \textcolor{DarkRed}{Oh-A-DINOv2} achieves a lower distance to the reference colour 
(\textbf{0.432}) compared to both \textcolor{DarkGoldenrod}{DINOv2} (\(0.512\)) and \textcolor{DarkGreen}{CLIP} (\(0.524\)). 
This indicates that \textcolor{DarkRed}{Oh-A-DINOv2} not only captures the global shape and orientation of cars but is also more sensitive 
to fine-grained attributes such as colour and material, which are crucial for reliable retrieval in real-world settings. 
The qualitative examples on the right confirm this behaviour: our method consistently retrieves cars that align in 
orientation and shape while matching the query’s visual appearance, addressing the key limitations of purely SSL-based features.

\section{Discussion}

\noindent\textbf{Limitations and Future Work.}
Our method depends on the quality of pre-trained features: if cues are absent, object-level latents may not suffice. Due to computational limitations, we did not deeply analyse why SSL representations neglect surface-level attributes; future work could study this or explore adjustments to pre-training. Beyond synthetic retrieval, testing on real-world control tasks where subtle attribute differences matter, and developing RL benchmarks that probe multi-object reasoning, would further validate the approach.

\noindent\textbf{Conclusion.}
We studied the ability of self-supervised and slot-based representations to preserve object attributes in multi-object scenes, showing consistent failures on surface-level cues such as colour and material.
To address this, we introduced a lightweight augmentation strategy: training a VAE on segmented patches from DINOv2 and combining its object-centric latents with the global embeddings.
This approach recovers missing attribute information while preserving the strengths of pre-trained models, yielding more faithful multi-object retrieval on synthetic and real-world datasets.
Our findings suggest that complementing large pre-trained representations with specialised object-centric latents is a promising direction for improving downstream tasks that require precise object-level understanding.

\bibliography{iclr2026_conference}
\bibliographystyle{iclr2026_conference}

\newpage
\section{Appendix}
\appendix
\section{Calculating similarities.} 
\label{sec:similarities}
We calculate the similarity on patch level between a query image $x$ and a candidate image $x'$ using the representations $v$ and $v'$. Since 
$v$ and $v'$ share the same embedding dimension, we can calculate the similarity between the two patches $v_i\in\mathbb{R}^{n_v}$ and $v'_j\in\mathbb{R}^{n_v}$ by computing the pairwise cosine similarity between the two representations:
\begin{equation}
    S_{ij} = \frac{v_i \cdot v'_j}{||v_i|| \cdot ||v'_j||} \in \mathbb{R},
\end{equation}
for all $i,j \in \{1, \ldots, p\}$.

Then, for each 
query patch $v_i$ we retrieve the patch with the highest cosine similarity, i.e., the most similar patch from $v'$
\begin{equation}
    s^{\text{max}}_i = \max_{j=\{1, \ldots, p\}} S_{ij}.
\end{equation}
The list $S^{\text{max}} = [s^{\text{max}}_1, \ldots, s^{\text{max}}_p]$ then contains the cosine similarities for the most similar patches from $v'$ for each patch in $v$.
Finally, our \emph{similarity score} of the query image $x$ with another image $x'$ is the average of the similarities in $S^{\text{max}}$, that is
$\bar{s} = \text{average}(S^{\text{max}})$. 
Note, if we compare $x$ against $n$ candidate images $x'$, we will have $n$ similarity scores for the given query image $x$.


\section{Learning Object-Level Features with PCA}
\subsection{The Benefit of Reapplying PCA}
\FloatBarrier
\begin{figure}[htbp!]
  \centering
  \includegraphics[scale=0.22]{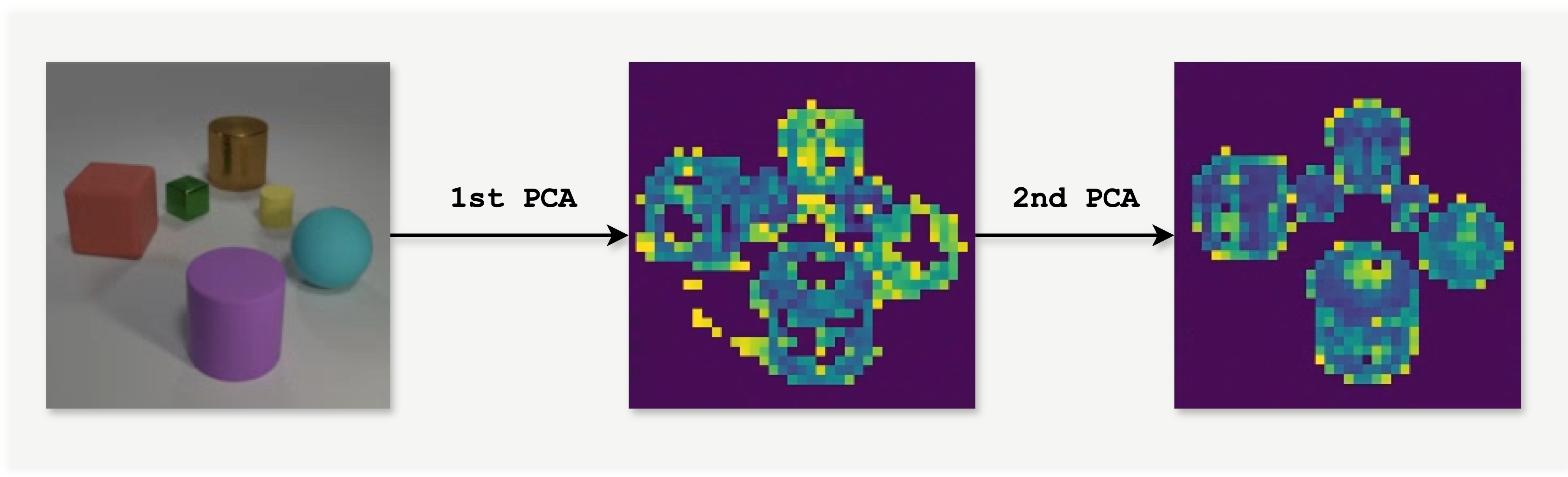}
  \caption{We show the benefit of reapplying PCA to the foreground object-level features. The first PCA passthrough to separate the background 
  and foreground features, while the second PCA pass-through helps to separate the individual objects and get a continuous mask across the objects.}
  \label{fig:2ndPCA}
\end{figure}
\FloatBarrier
In Figure \ref{fig:2ndPCA}, we show the benefit of applying PCA a second time, i.e., over the segmented foreground features after separating background from foreground. 
While the first PCA provides a decent mask, some noisy patches from the background still remain and some patches on the objects are segmented away. 
Applying PCA a second time on the foreground patch features which were separated from the background in the first pass, helps in retrieving a better overall
segmentation mask.
\section{Additional Dataset Details}
\label{sec:additional}
\subsection{Dataset Statistics}
\FloatBarrier
\begin{table}[htbp!]
  \centering
  {\fontfamily{pcr}\selectfont 
  \scalebox{0.8}{
  \begin{tabular}{|c|c|c|c|c|c|c|}
  \cline{1-7} 
  \emph{Dataset} & \emph{\#Images} & \emph{\#Objects}& \emph{\#Shapes} & \emph{\#Colors} & \emph{\#Materials} & \emph{\#Backgrounds}\\ \cline{1-7}
  \textbf{CLEVR} & 100k & 3-10& 3 & 8 & 2 & 1 \\ \cline{1-7}
  \textbf{CLEVRTex} & 50k+10k & 3-10 & 4& - & 60 & 60  \\ \cline{1-7}   
  \end{tabular}}}
  \vspace{2ex}
  \caption{Dataset characteristics for CLEVR and CLEVRTex. The CLEVRTex dataset has additional textures for the objects and backgrounds compared to CLEVR. It also has no color information for the objects, which is replaced by the material information.
  In general, instance retrieval with the CLEVRTex dataset is much harder due to the increased number of possible object configurations.}
  \label{tab:dataset_char}
\end{table}
We provide details on the number of attributes and objects in the CLEVR and CLEVRTex dataset in Table \ref{tab:dataset_char}. While both 
datasets have the same number of possible objects in an image, the CLEVRTex Dataset has different materials compared to the CLEVR dataset 
which has 8 colors and 2 materials. This presents different challenges to the learnt reprsentations. In the CLEVR case, the representations must account 
for redundancy in the object features, while for the CLEVRTex dataset the large number of materials makes retrieval more challenging since some materials 
are quite similar and only differ slightly.
\FloatBarrier
\subsection{Training-Test Set Sizes}
\FloatBarrier
\begin{table}[htbp!]
    \centering
    {\fontfamily{pcr}\selectfont \footnotesize
    \scalebox{1.0}{
    \begin{tabular}{|c|c|c|c|}
    \multicolumn{4}{c}{\textbf{CLEVR/CLEVRTex Dataset Splits}}\\ \cline{1-4}
    \cline{1-4} 
    \emph{Train} & \emph{Validation-Query} & \emph{Candidates}& \emph{Query Size}\\ \cline{1-4}
    2000 & 500 & 5000 & 50 \\ \cline{1-4}
    \end{tabular}}}
    \vspace{2ex}
    \caption{\textbf{Data splits for the CLEVR and CLEVRTex datasets.} The train set is used to train the VAE, while the validation set is used to select query images for the retrieval task. The test set contains the candidate images for the retrieval task.}
    \label{tab:splits}
\end{table}
\FloatBarrier

\FloatBarrier
\begin{table}[htbp!]
    \centering
    {\fontfamily{pcr}\selectfont \footnotesize
    \scalebox{1.0}{
    \begin{tabular}{|c|c|c|}
    \multicolumn{3}{c}{\textbf{Stanford Cars Dataset Splits}}\\ \cline{1-3}
    \cline{1-3} 
    \emph{Validation-Query} & \emph{Candidates}& \emph{Query Size}\\ \cline{1-3}
     175 & 5000 & 25 \\ \cline{1-3}
    \end{tabular}}}
    \vspace{2ex}
    \caption{\textbf{Data splits for Stanford Cars dataset.} We only evaluate on the Stanford Cars Dataset where we take a subset of 5000 candidate images and 175 query images split into groups of 25.}
    \label{tab:splits}
\end{table}
\FloatBarrier

\section{Experiment Details}
\label{sec:hyperparameters}
\subsection{VAE Hyperparameters}
\FloatBarrier
\begin{table}[htbp!]
  \centering
  {\fontfamily{pcr}\selectfont \footnotesize
  \scalebox{1.0}{
  \begin{tabular}{|c|c|}
  \multicolumn{2}{c}{\textbf{VAE Hyperparameters}}\\ \cline{1-2}
  \cline{1-2} 
  \emph{Parameter} & \emph{Value} \\ \cline{1-2}
  Input size & $64 \times 64 \times 3$  \\ \cline{1-2}
  Latent size & 32  \\ \cline{1-2}
  Learning rate & $1e-4$  \\ \cline{1-2}
  $\beta$ & $1e-4$ \\ \cline{1-2}

  \end{tabular}}}
\end{table}
\FloatBarrier
\subsection{DINO Hyperparameters}
\FloatBarrier
\begin{table}[htbp!]
  \centering
  {\fontfamily{pcr}\selectfont \footnotesize
  \scalebox{1.0}{
  \begin{tabular}{|c|c|}
  \multicolumn{2}{c}{\textbf{DINO Hyperparameters}}\\ \cline{1-2}
  \cline{1-2} 
  \emph{Parameter} & \emph{Value} \\ \cline{1-2}
  Input size & $518 \times 518$  \\ \cline{1-2}
  Embedding Dimension & $384$  \\ \cline{1-2}
  Patch Size & $14$  \\ \cline{1-2}
  \end{tabular}}}
\end{table}
\FloatBarrier
Note that for DINOv2 and DINOv3 we used the non register variants since we noticed no difference in our analysis and the non register variants were faster.
\subsection{Compute}
The experiments where performed on an RTX 3090 24GB VRAM with Ryzen 3900X CPU. The main bottleneck is running the experiments which requires 
caching all the required query and candidate images. For this, we used 128GB of RAM.

\section{Additional Visualisations}
\label{sec:vis}
\subsection{CLEVRTex}
We show $3$ retrievals from the query dataset for CLEVRTex where relative performance 
is comparable, i.e., the retrievals are ranked similar in terms of retrieval performance within each model. 
We see that Oh-A-DINO, performs the best retrievals, in most cases being able to retrieve the exact object 
or an object similar to the reference object. DINOv2 as presented in Figure \ref{fig:intro}  retrieves the correct 
object, however with the wrong texture properties. For the VAE retrievals, we see that the object properties are understood correctly,
however with no real object attribute binding, often retrieving the wrong objects or the background in the material of the object. 
This shows, that the combination of the VAE representation an the DINOv2 representation allows for correctly interpreting the background
while performing consistent object retrieval. SlotDiffusion performs the worst, being very sensitive to the background. Even when masking the background
with our approach, results barely improved. 
\FloatBarrier
\begin{figure}[htbp!]
  \centering
  \includegraphics[scale=0.35]{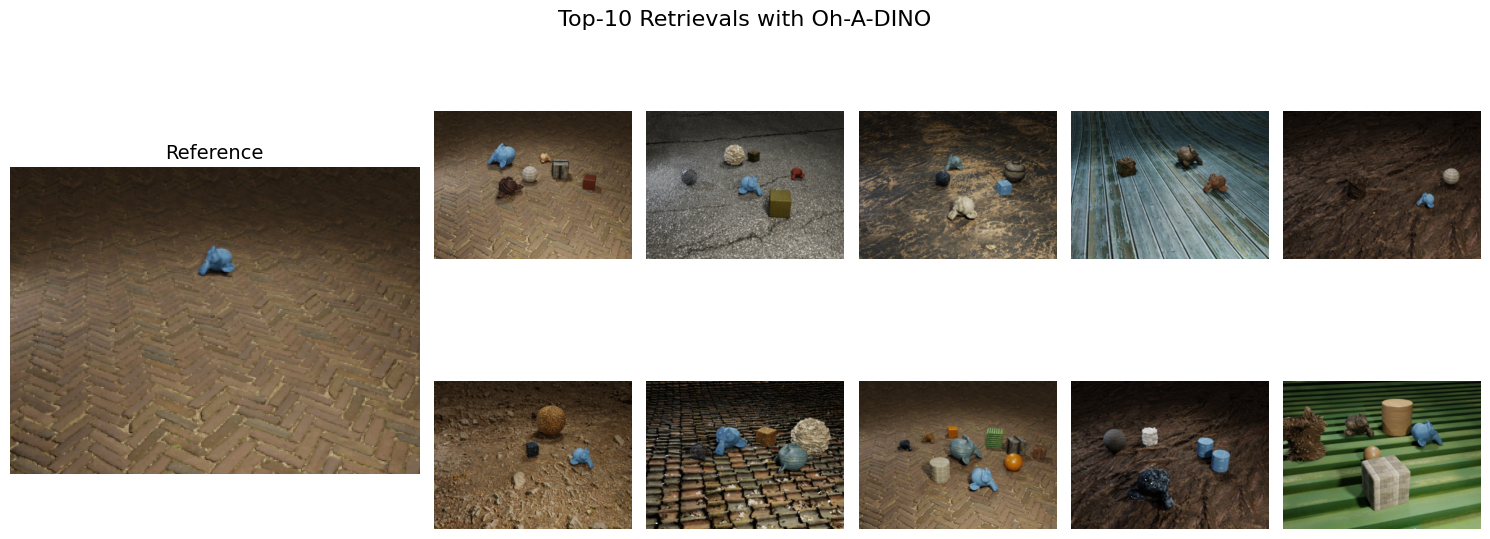}
\end{figure}
\FloatBarrier
\FloatBarrier
\begin{figure}[htbp!]
  \centering
  \includegraphics[scale=0.35]{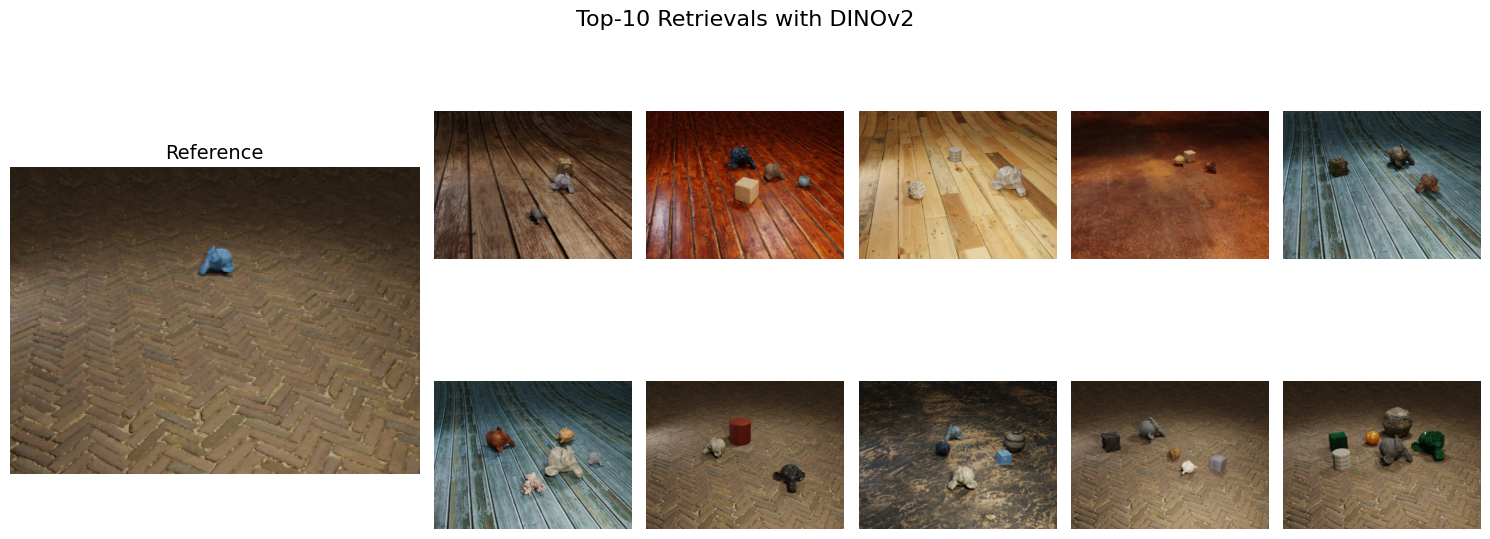}
\end{figure}
\FloatBarrier
\FloatBarrier
\begin{figure}[htbp!]
  \centering
  \includegraphics[scale=0.35]{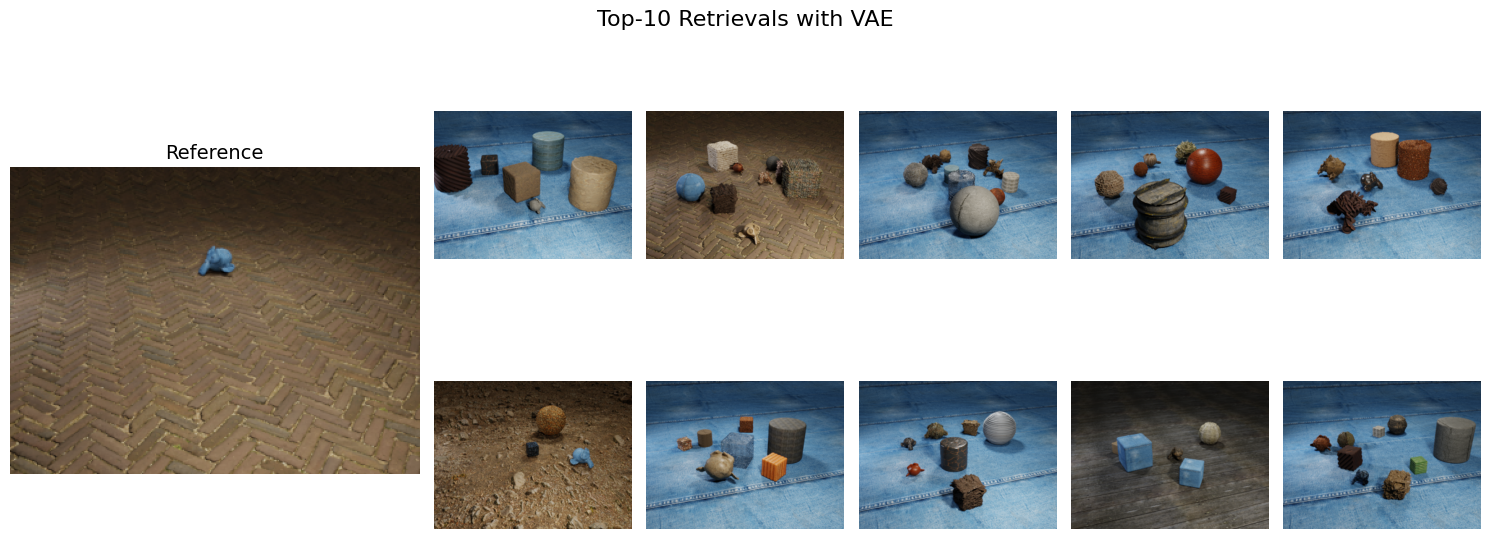}
\end{figure}
\FloatBarrier
\FloatBarrier
\begin{figure}[htbp!]
  \centering
  \includegraphics[scale=0.35]{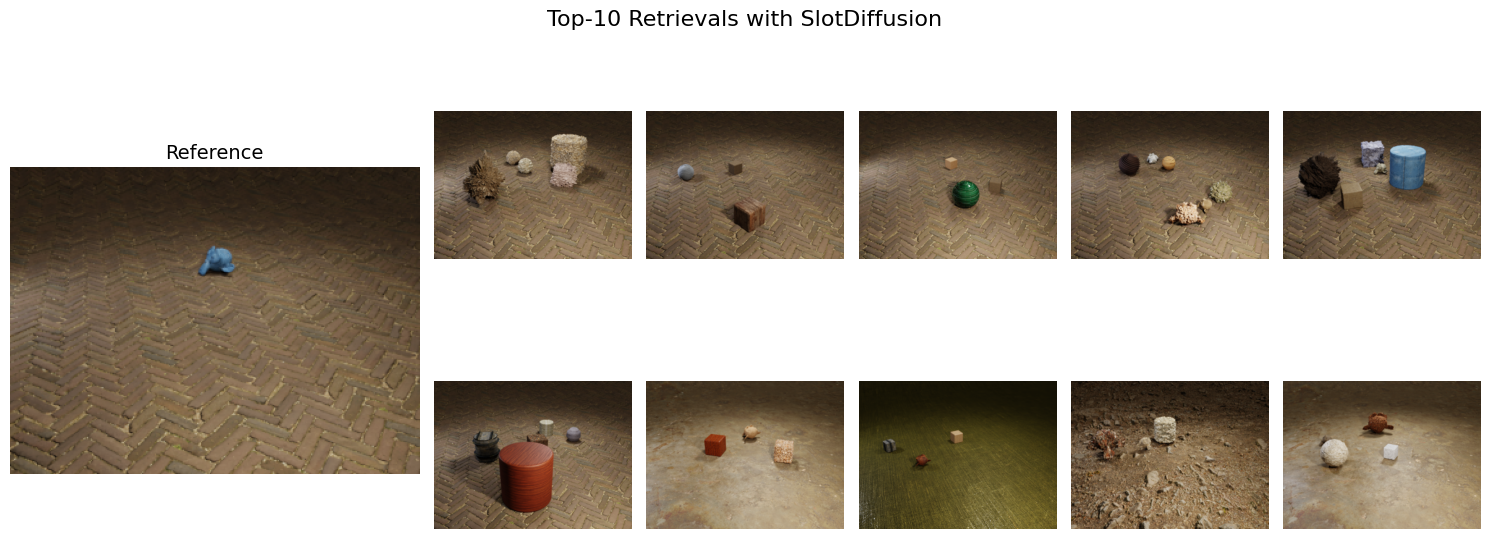}
\end{figure}
\FloatBarrier
\FloatBarrier
\begin{figure}[htbp!]
  \centering
  \includegraphics[scale=0.35]{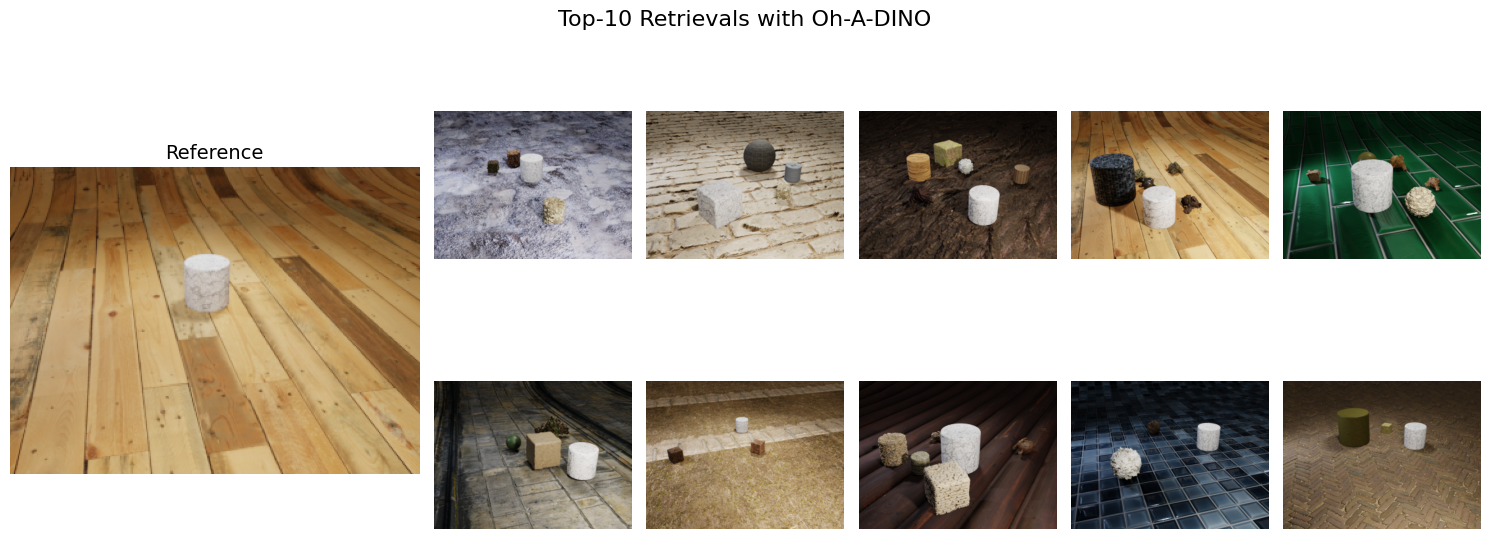}
\end{figure}
\FloatBarrier
\FloatBarrier
\begin{figure}[htbp!]
  \centering
  \includegraphics[scale=0.35]{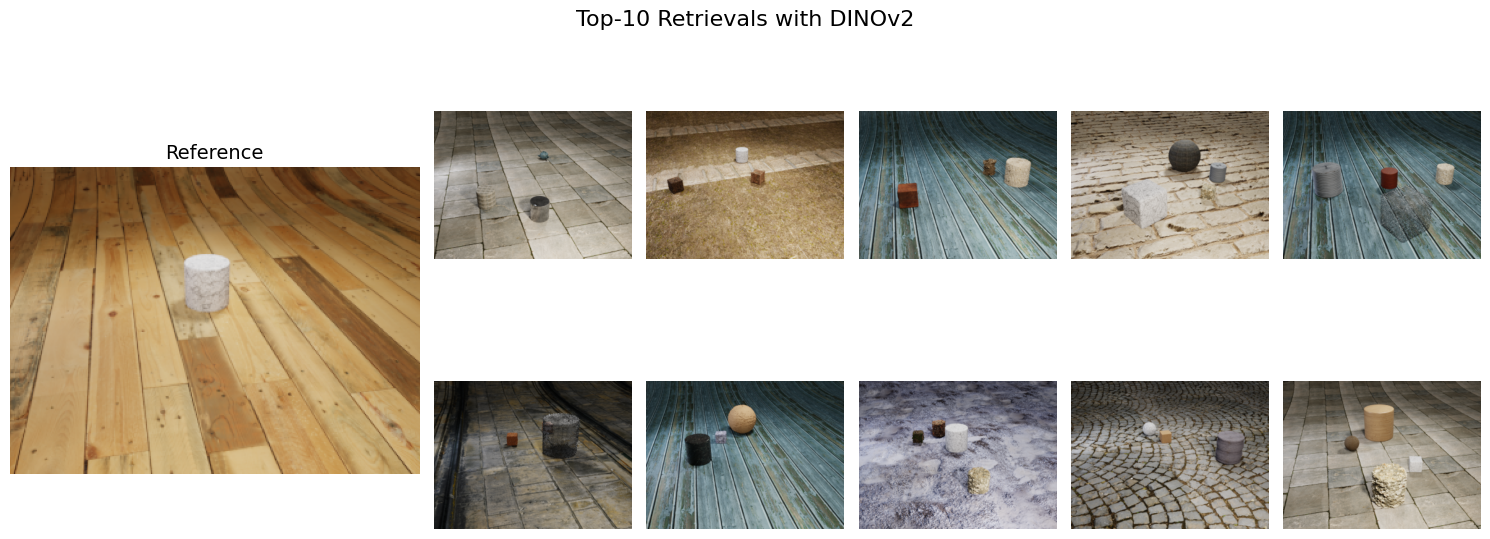}
\end{figure}
\FloatBarrier
\FloatBarrier
\begin{figure}[htbp!]
  \centering
  \includegraphics[scale=0.35]{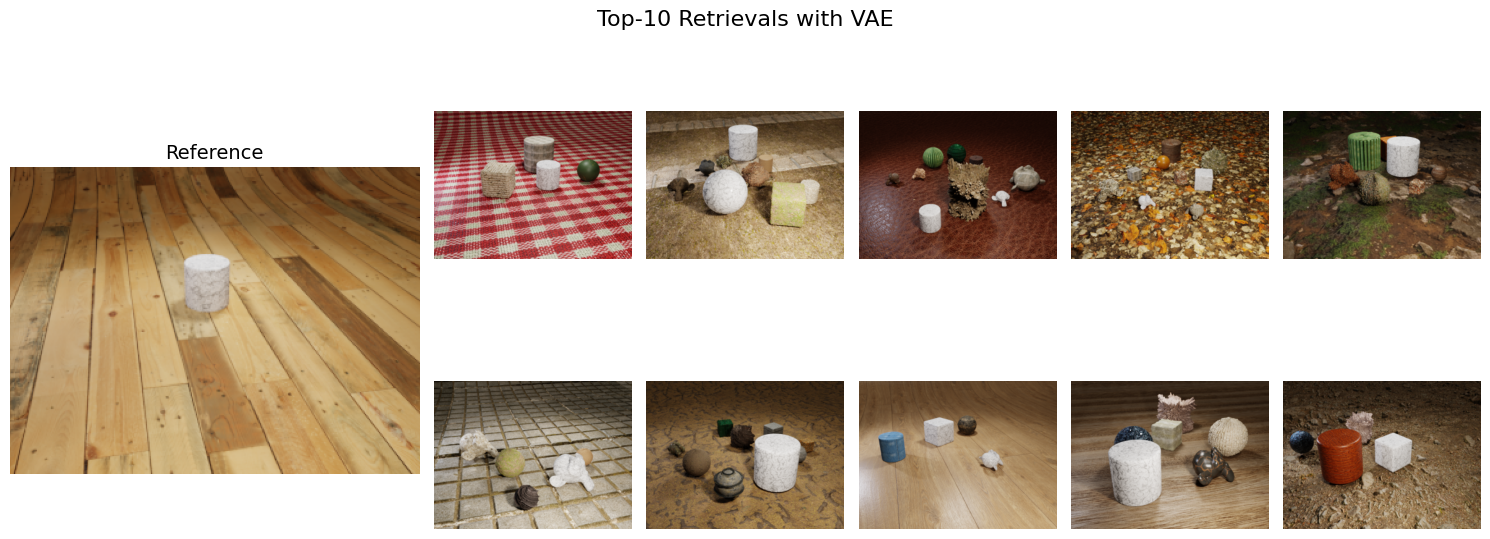}
\end{figure}
\begin{figure}[htbp!]
    \centering
    \includegraphics[scale=0.35]{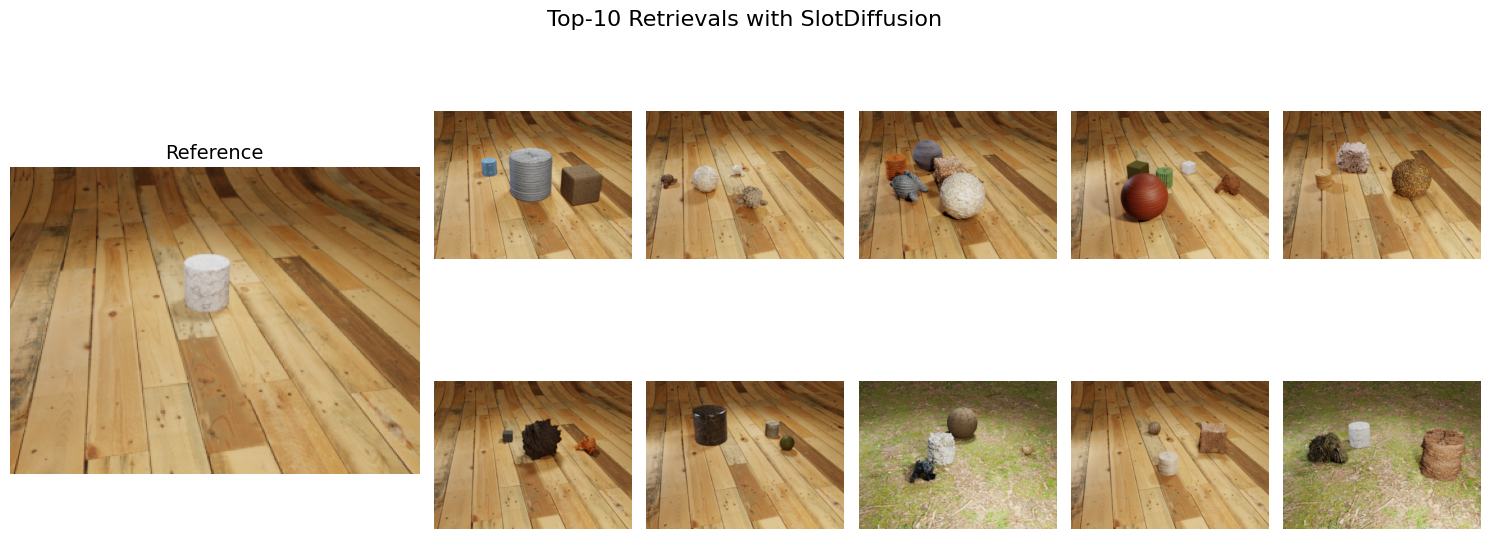}
  \end{figure}
\FloatBarrier
\FloatBarrier
\begin{figure}[htbp!]
  \centering
  \includegraphics[scale=0.35]{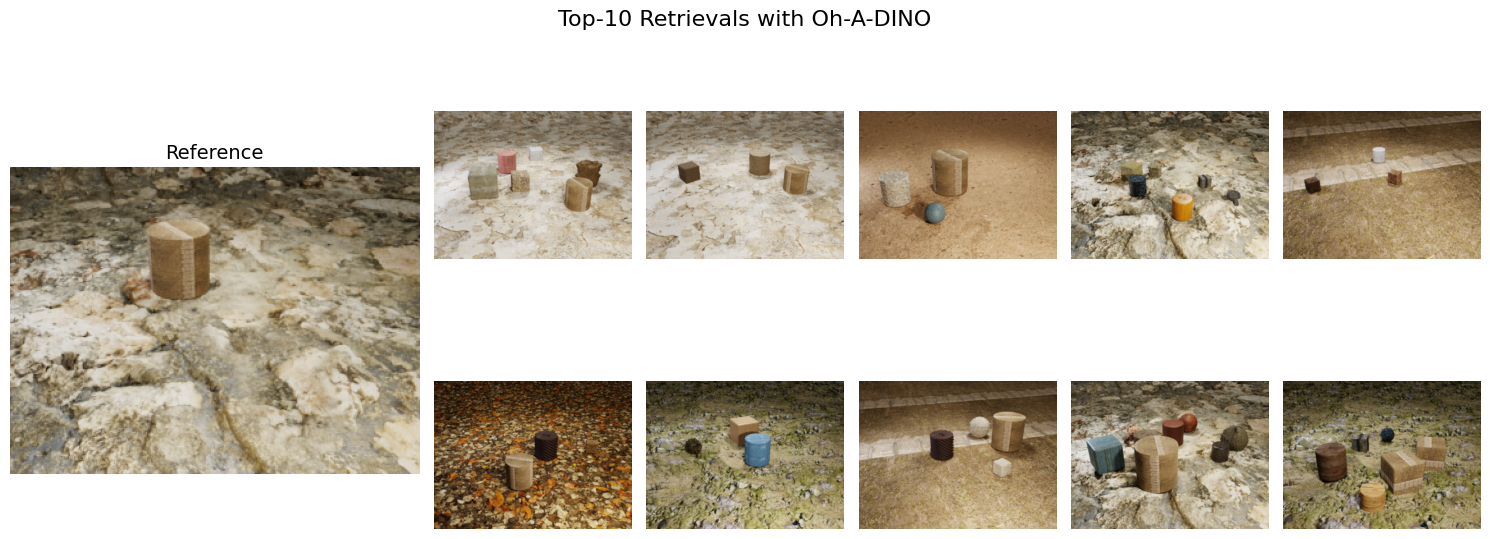}
\end{figure}
\FloatBarrier
\FloatBarrier
\begin{figure}[htbp!]
  \centering
  \includegraphics[scale=0.35]{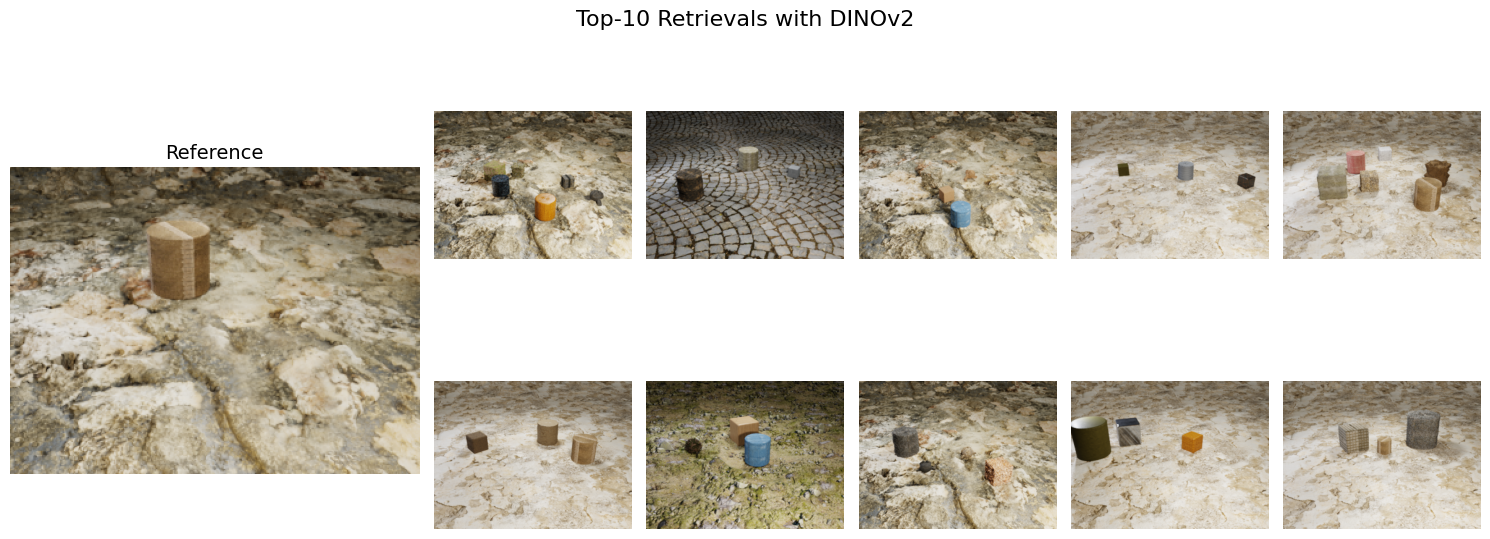}
\end{figure}
\FloatBarrier
\FloatBarrier
\begin{figure}[htbp!]
  \centering
  \includegraphics[scale=0.35]{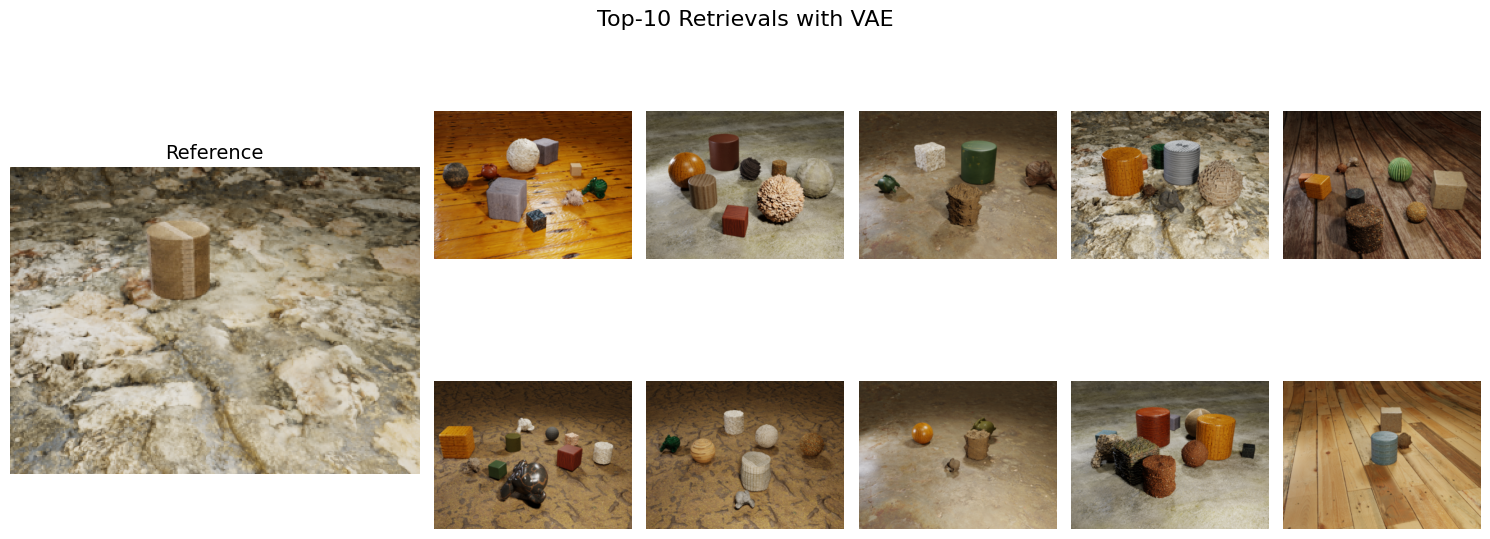}
\end{figure}
\begin{figure}[htbp!]
    \centering
    \includegraphics[scale=0.35]{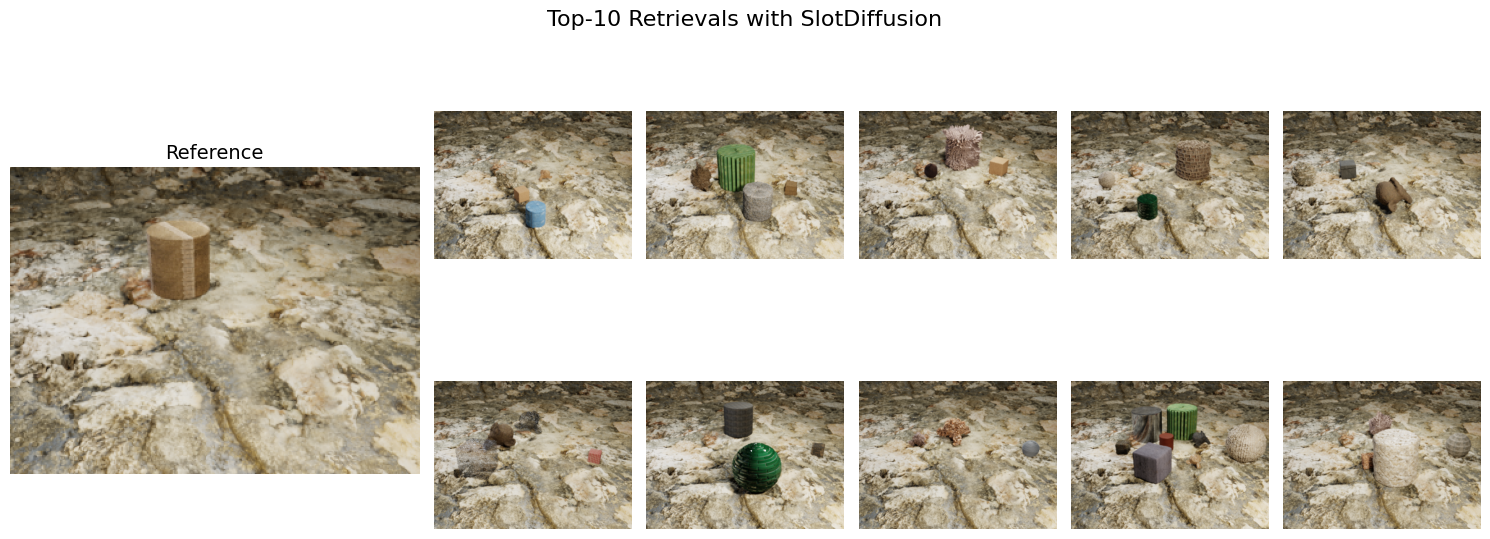}
  \end{figure}
\FloatBarrier

\subsection{CLEVR}
Compared to CLEVRTex the differences in retrieval performance are more nuanced. We see that Oh-A-DINO is able to retrieve all objects almost 
without errors, compared to the DINOv2 representations, which again retrieves the correct objects, but with wrong color or sometimes object properties. 
The VAE object-level features perform well in CLEVR due to the lack of distractors such as the background, however when comparing to the Oh-A-DINO retrievals, 
we observe that the retrievals with VAE respect the scene structure less. For instance, at times the object in the retrieval is not in the same position, orientation
as the reference object or the same object is present multiple times in the scene. Again, SlotDiffusion performs the worst. Compared to CLEVRTex we can see 
that the SlotDiffusion representations retrieve similar objects, but often fails to bind different attributes together resulting in difficult to interpret retrievals, confirming
our numerical results.
\FloatBarrier
\begin{figure}[htbp!]
  \centering
  \includegraphics[scale=0.35]{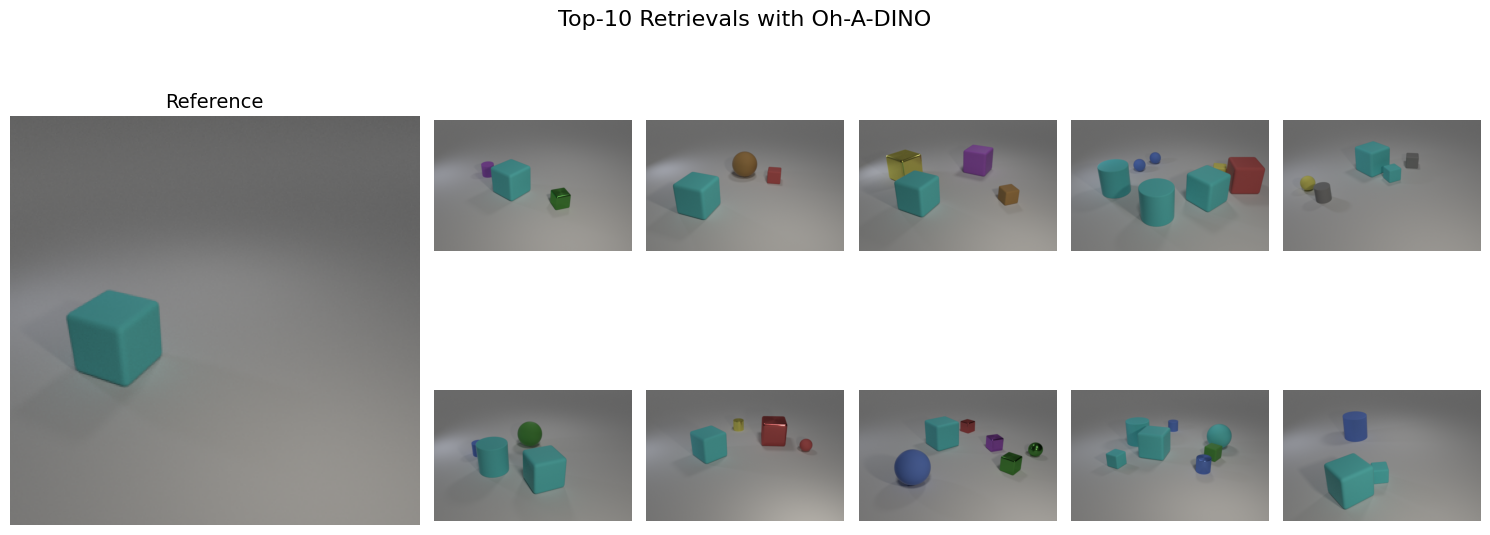}
\end{figure}
\FloatBarrier
\FloatBarrier
\begin{figure}[htbp!]
  \centering
  \includegraphics[scale=0.35]{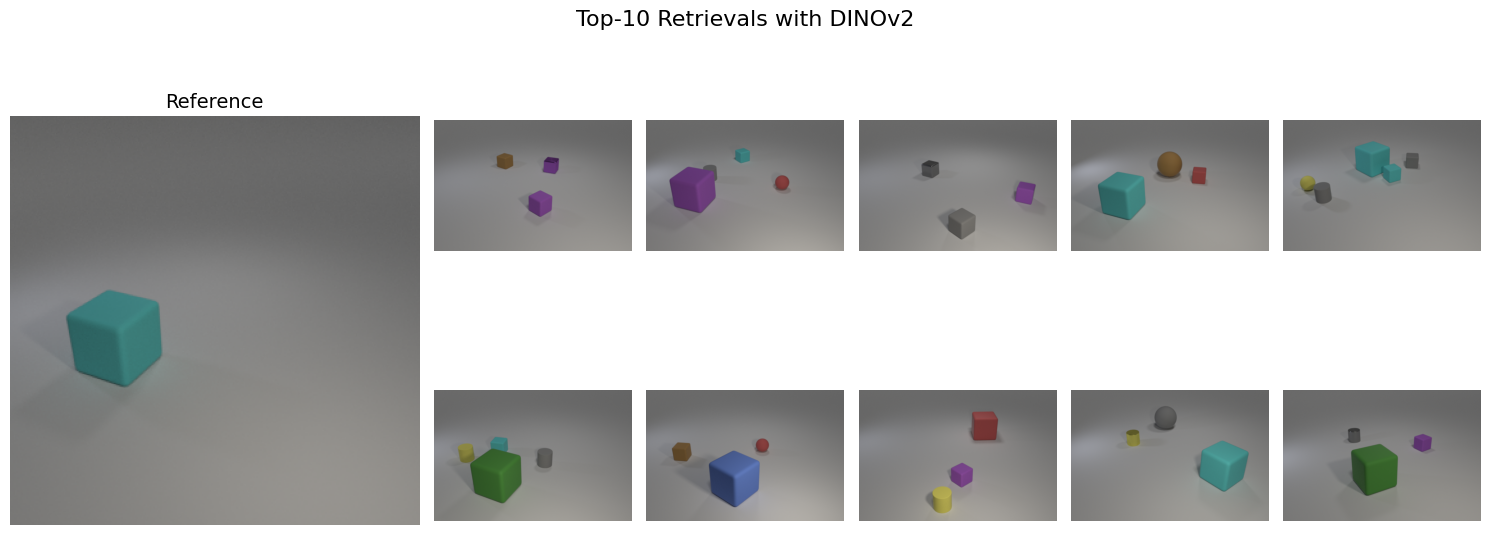}
\end{figure}
\FloatBarrier
\FloatBarrier
\begin{figure}[htbp!]
  \centering
  \includegraphics[scale=0.35]{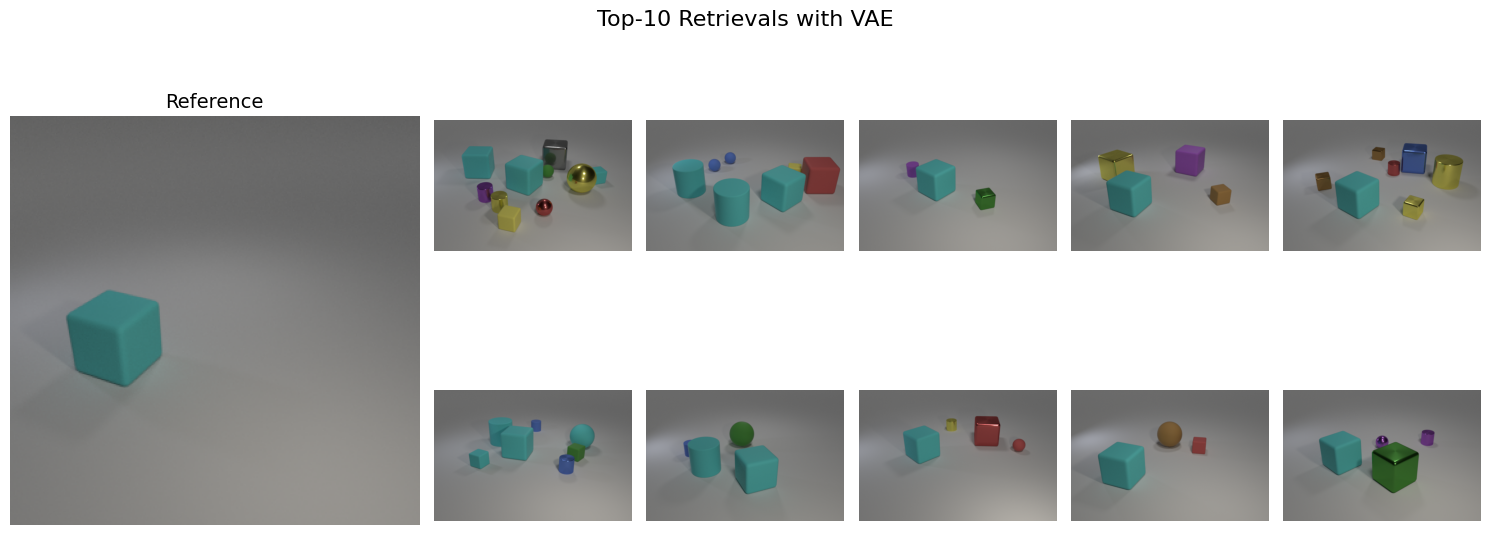}
\end{figure}
\FloatBarrier
\FloatBarrier
\begin{figure}[htbp!]
  \centering
  \includegraphics[scale=0.35]{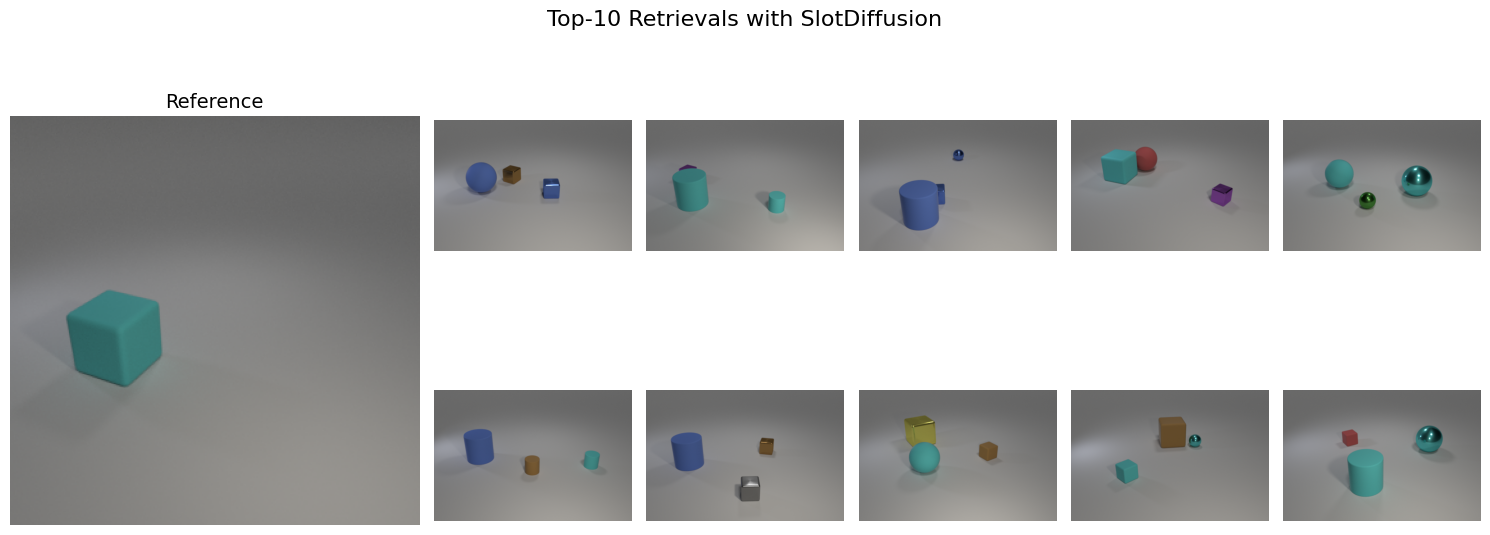}
\end{figure}
\FloatBarrier
\FloatBarrier
\begin{figure}[htbp!]
  \centering
  \includegraphics[scale=0.35]{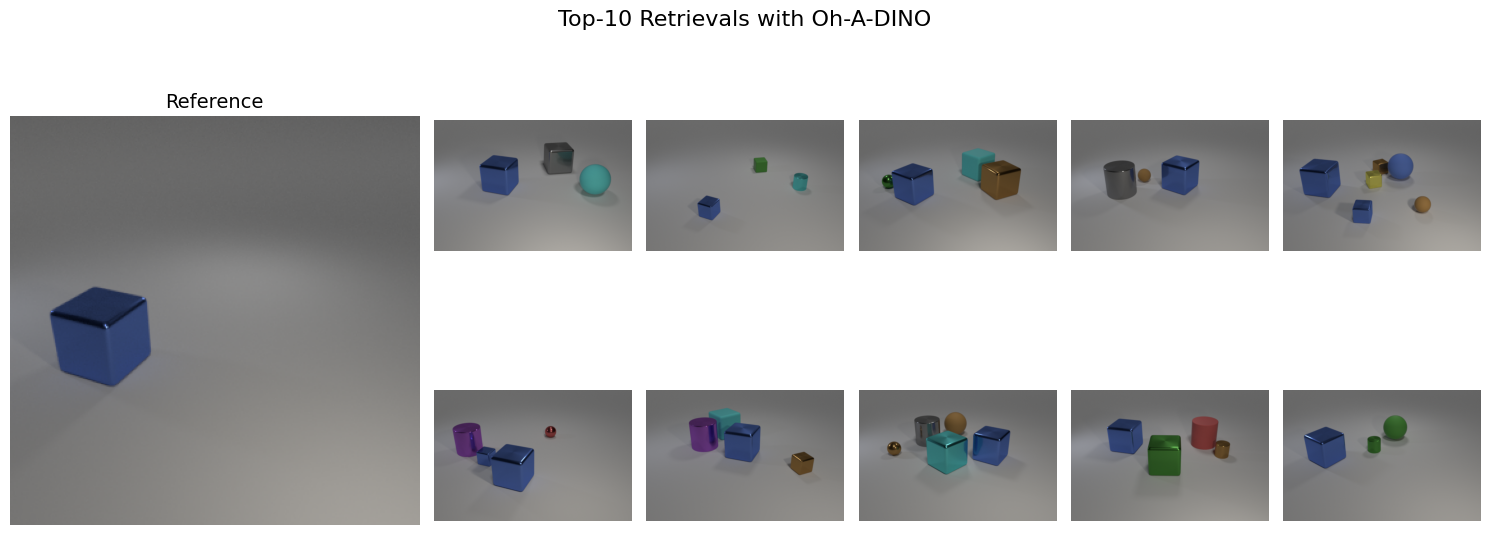}
\end{figure}
\FloatBarrier
\FloatBarrier
\begin{figure}[htbp!]
  \centering
  \includegraphics[scale=0.35]{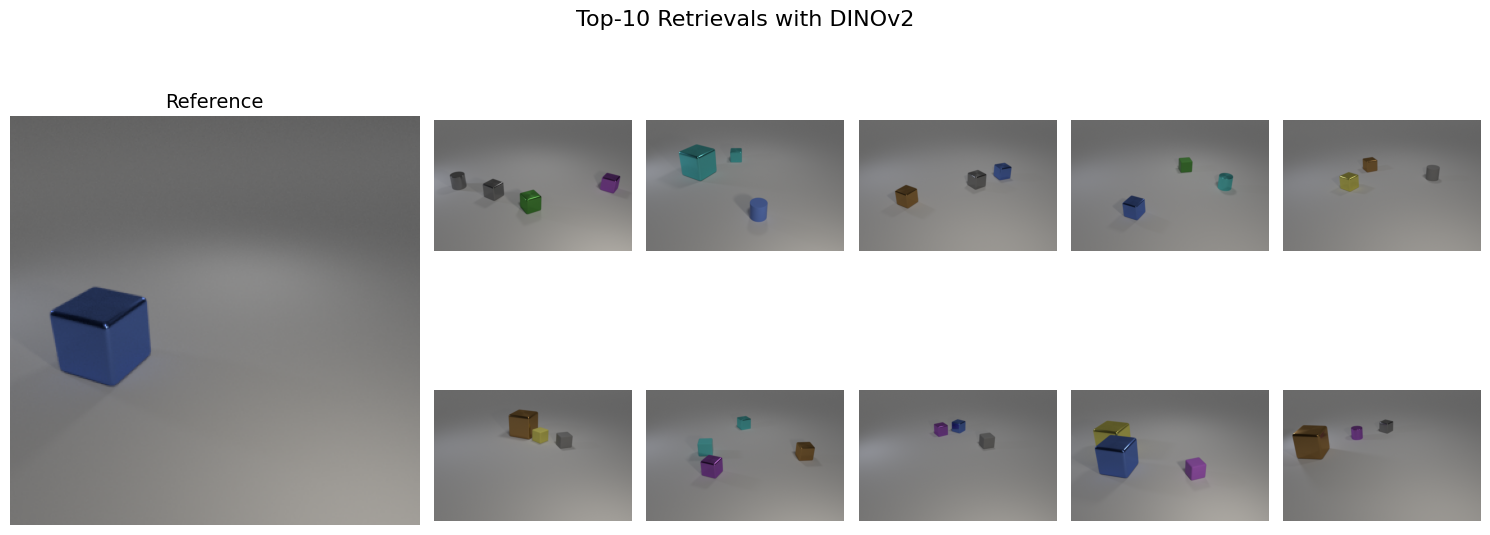}
\end{figure}
\FloatBarrier
\FloatBarrier
\begin{figure}[htbp!]
  \centering
  \includegraphics[scale=0.35]{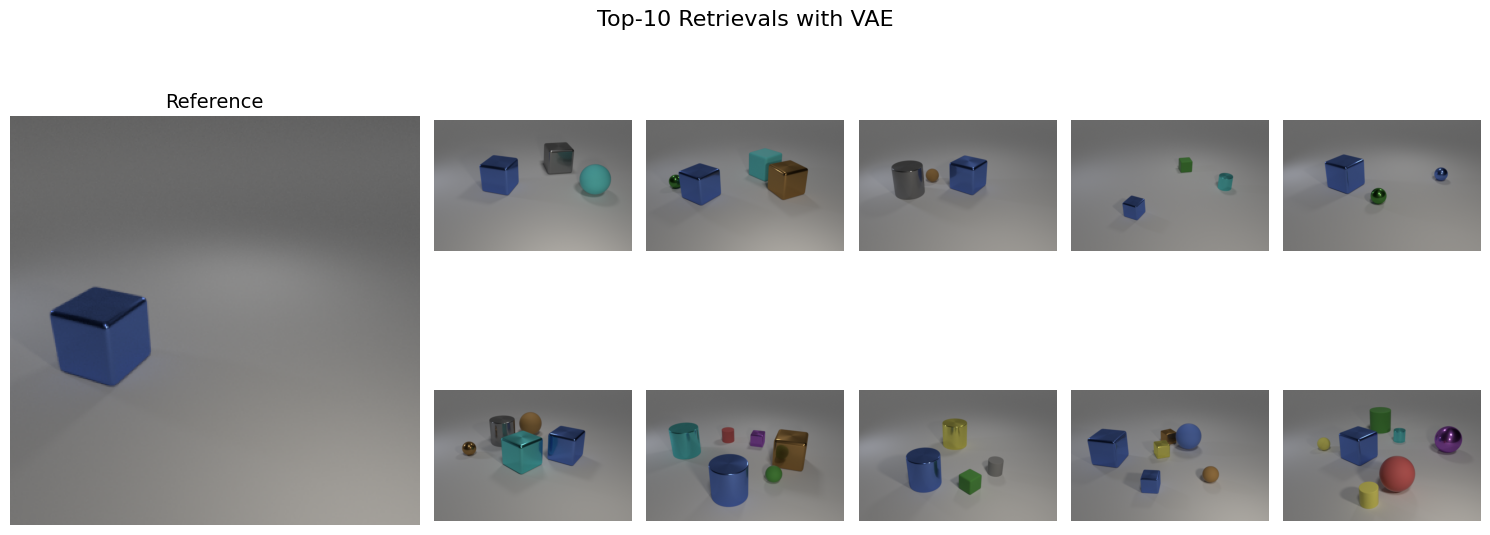}
\end{figure}
\begin{figure}[htbp!]
    \centering
    \includegraphics[scale=0.35]{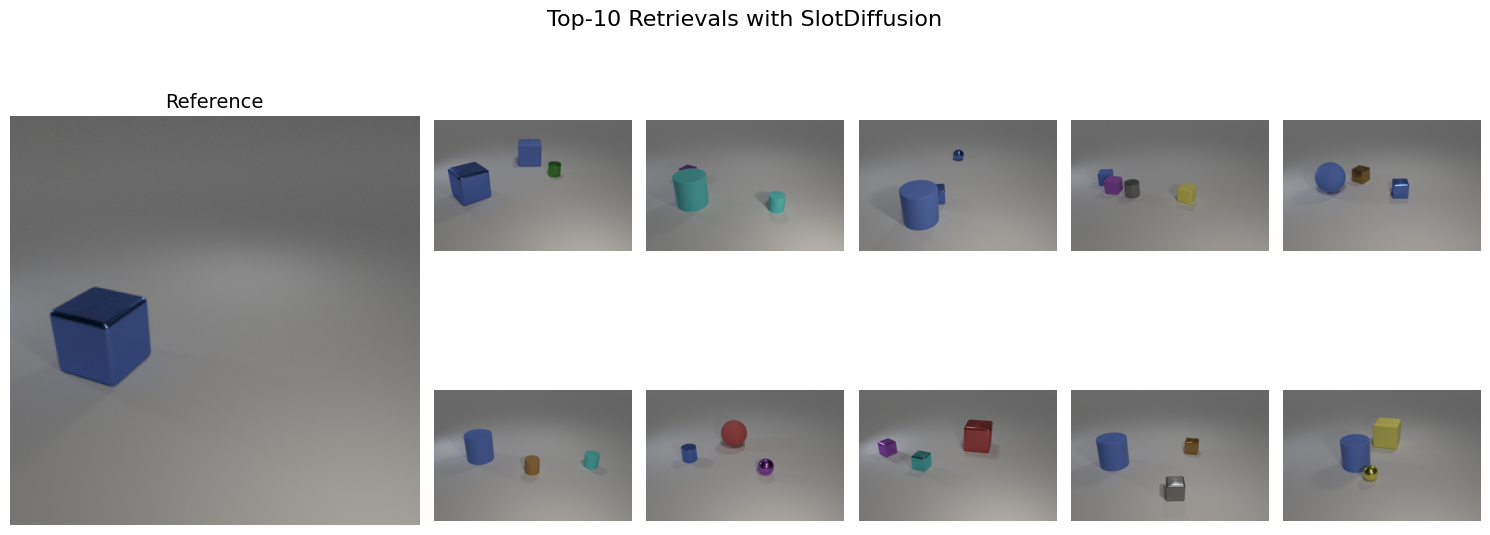}
  \end{figure}
\FloatBarrier
\FloatBarrier
\begin{figure}[htbp!]
  \centering
  \includegraphics[scale=0.35]{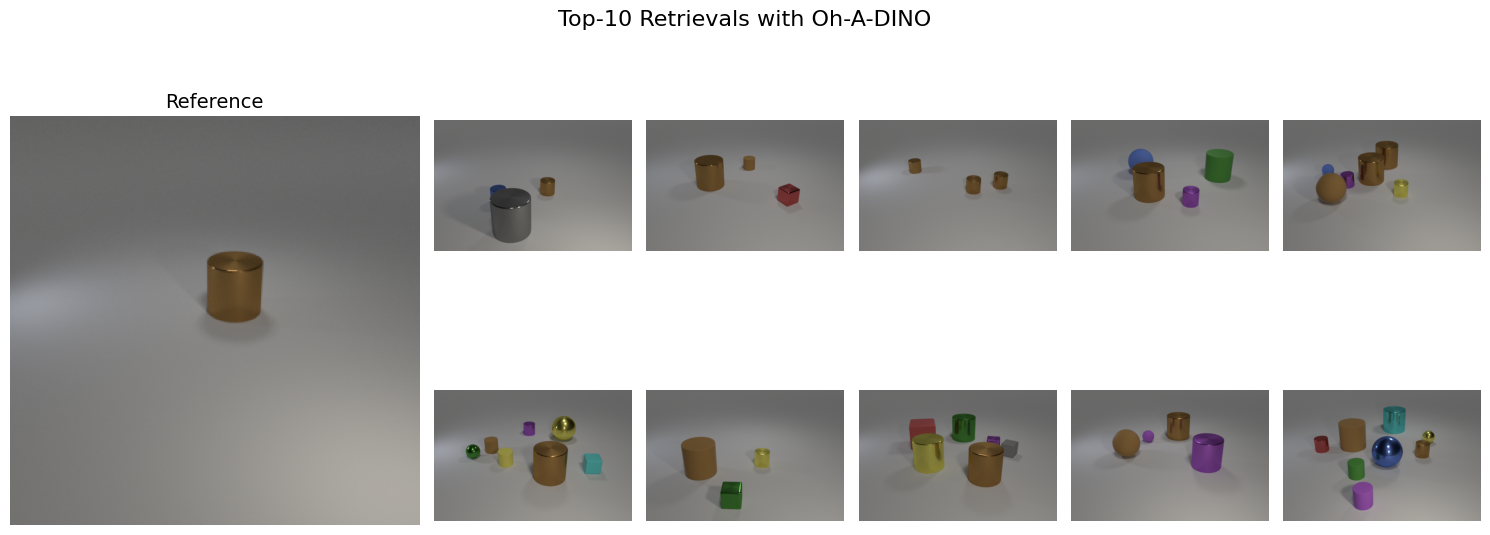}
\end{figure}
\FloatBarrier
\FloatBarrier
\begin{figure}[htbp!]
  \centering
  \includegraphics[scale=0.35]{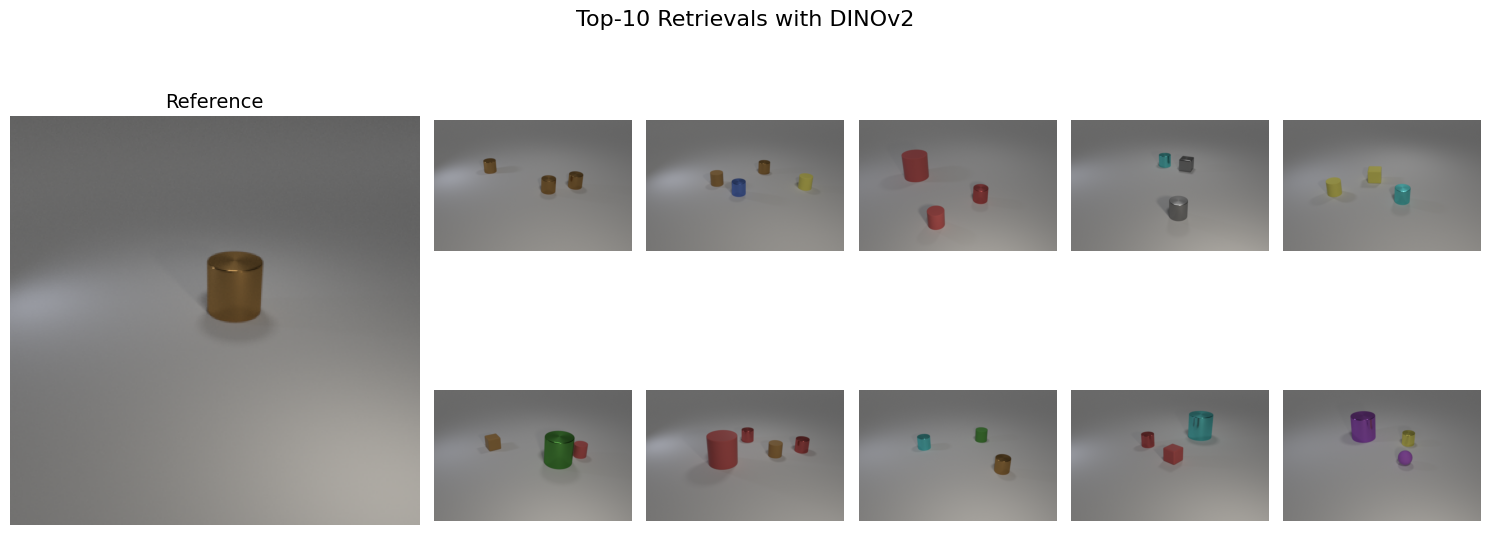}
\end{figure}
\FloatBarrier
\FloatBarrier
\begin{figure}[htbp!]
  \centering
  \includegraphics[scale=0.35]{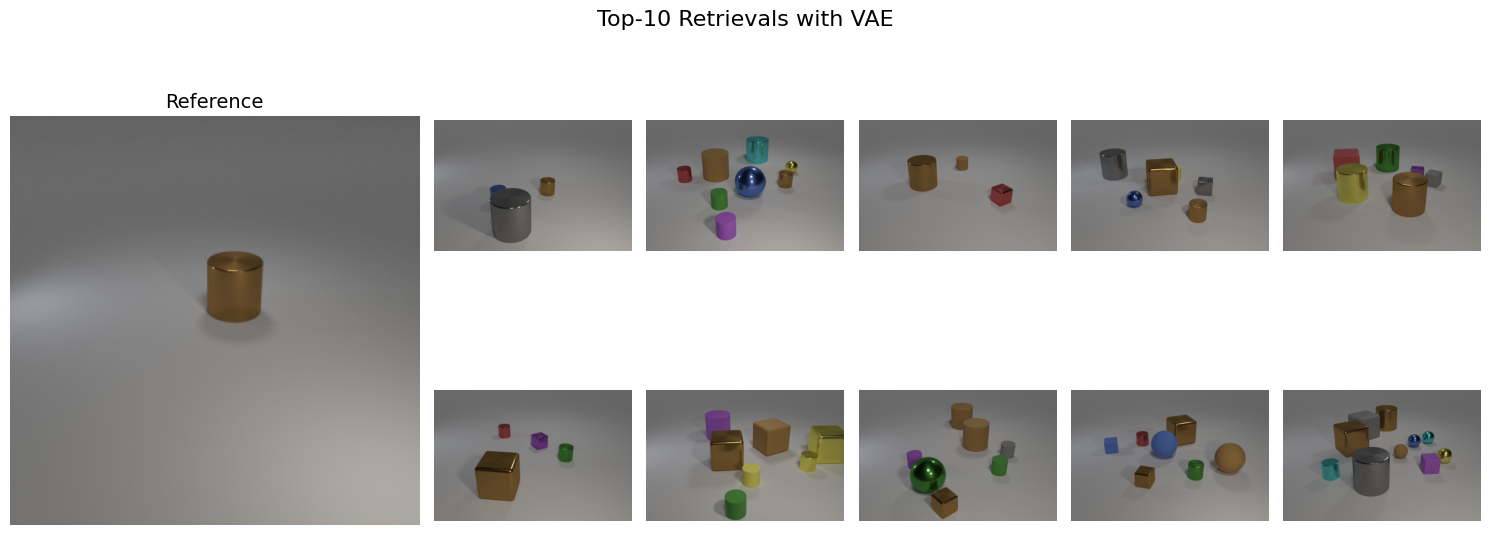}
\end{figure}
\begin{figure}[htbp!]
    \centering
    \includegraphics[scale=0.35]{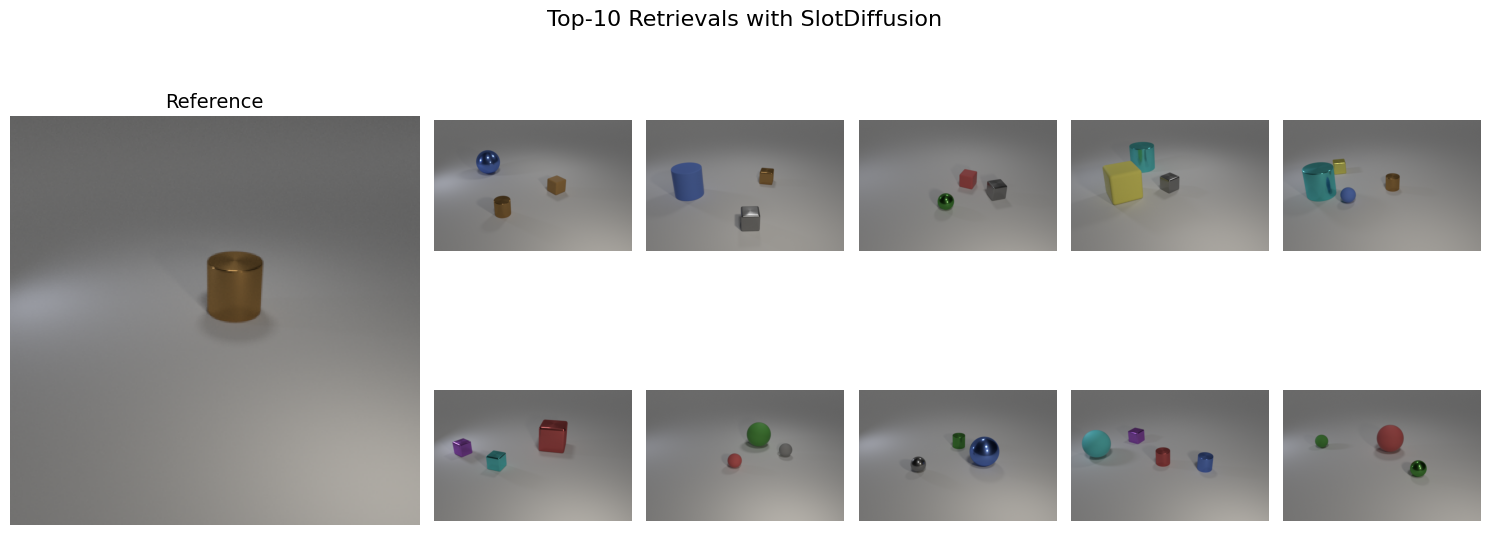}
  \end{figure}
\FloatBarrier
\end{document}